# Evidence sufficiency assessment enables cost-effective review allocation in on-premise clinical AI agents

**Authors**: Yiwen Qiu[1,†], Che Sun[2,†], Shu Shen[1], Wentao Wang[1,*]

**Affiliations**:

[1] Department of Liver Surgery, West China Hospital of Sichuan University, Chengdu, China

[2] Shenzhen Mindray Bio-Medical Electronics Co., Ltd., Shenzhen, China

†These authors contributed equally: Yiwen Qiu, Che Sun (Yiwen Qiu is listed first).

**Corresponding author**: * Wentao Wang

**Email**: wwtdoctor02@163.com

## Abstract

On-premise clinical AI agents must auto-accept high-confidence diagnoses, deferring others to clinician review, yet token probabilities and self-reported confidence miss errors, while more reliable five-run consistency incurs fivefold compute. We propose training-free decision-point evidence sufficiency assessment and a three-tier sequential triage protocol. Masking reasoning and diagnosis, the same 4B model scores sufficiency from objective evidence via prior-debiased probabilities over five ordered levels. Calibration freezes a nominal top-50% threshold, selecting single-run, five-run consistency, or full-deferral tiers by whether the one-sided 95% Wilson lower confidence bound of pass rate among accepted cases meets the 95% clinical target line. Across 1,605 MIRA-v2, CDM, VivaBench, and in-hospital liver malignancy primary cohort cases, MIRA-v2 single-run top-50% pass rate reached 98.91%, matching five-run consistency, saving ~80% compute; CDM achieved 96.00% test attainment by escalating to the five-run consistency tier; cross-specialty, missing-imaging, and combined enriched cohorts entered the full-deferral tier, capping calibration false-certification rate at 0.00%–3.50%.

## Introduction

Large language model-driven clinical AI agents have evolved into multi-turn diagnostic systems capable of taking medical histories, ordering laboratory tests and imaging studies, and formulating diagnoses[1], moving beyond earlier single-turn medical question answering and static disease diagnosis[2,3]. However, integrating such systems into routine clinical care faces a dual constraint: on one hand, patient privacy regulations such as the US Health Insurance Portability and Accountability Act (HIPAA; https://www.hhs.gov/hipaa/) and the European Health Data Space Regulation[4], along with on-premise data-residency and accountability auditing requirements, often motivate hospitals to deploy models within compute-constrained local environments[5], where operational concurrency typically limits inference to a single consultation per case; on the other hand, prior experience with clinical decision support systems shows that clinicians under heavy workloads cannot inspect every long-horizon consultation trajectory, and mandatory case-by-case verification induces "alert fatigue" that leads to reflexive dismissal or habitual overriding[6,7]. Hospitals therefore urgently need reliability signals to operationalize selective autonomy at the diagnostic submission point through a triage workflow that automatically accepts high-confidence cases and defers the remainder to clinician review[8,9]. Because early omissions in history taking or biases in ordering investigations propagate and amplify across turns, assessing the reliability of diagnostic outputs from multi-turn clinical AI agents is substantially more challenging than single-turn question answering[10].

Recent work has systematically compared three classes of reliability signals for on-premise clinical AI agents[11], yet existing approaches still struggle to reconcile the low compute overhead of a single consultation with robustness across clinical conditions. First, intrinsic probability or self-reported signals extracted from a single consultation suffer from weak discrimination or unstable triage thresholds: token generation probability primarily reflects local linguistic fluency or model preference rather than diagnostic correctness[11] and can be further diluted by routine documentation phrasing in long consultations, while self-reported diagnostic confidence is widely miscalibrated[12,13] and frequently exhibits heavy tie clustering at high scores such as 1.0 and 0.95, preventing hospitals from uniquely determining a triage threshold under a fixed auto-accept fraction. Second, cross-run consistency methods such as self-consistency and semantic entropy[14–16] achieve the strongest discrimination among on-premise clinical AI agent comparators, yet require repeated sampling that typically incurs five times the inference computation in multi-turn evaluations[11]. Third, structured decision readout methods read the output logits over a finite set of ordered options and apply prior debiasing without generating free text[17]. Nevertheless, prior medical explorations have largely been confined to static single-step interpretation, relied on large-scale fine-tuning, or evaluated only single-turn triage[18–21], remaining vulnerable in clinical use to option-naming effects[22,23], ordinal-scale bias[24], and insufficient abstention when the correct option or key clinical evidence is missing[22,25].

No single reliability signal generalizes across heterogeneous clinical cohorts, so hospitals require a pre-deployment admission framework capable of both low-cost auto-acceptance and timely cessation of autonomous handling. Although prior studies have proposed a general statistical certification paradigm that sets thresholds on a local calibration set before verifying them against a lower confidence bound on an independent held-out set[26], alongside strategies such as selective abstention, model cascading, and adaptive sampling[27–31], most were designed for single-turn static text or cloud-scale models and lack a triage protocol suited to on-premise multi-turn consultations. When extended across clinical departments and disease spectra, triage decisions face three characteristic failure risks: in syndromes with substantially overlapping clinical phenotypes such as acute abdominal pain, the pass rate among accepted cases from a single consultation may lie only slightly above the minimum clinically acceptable standard and fall short under sampling variation; in cross-specialty open consultations, the differential capability of lightweight local models may itself be insufficient to support safe auto-acceptance at any fraction; and in real-world specialty records, missing key imaging conclusions or enrichment of atypical pathological subtypes elevates the intrinsic misdiagnosis rate among high-scoring cases, causing repeated runs to reproduce identical wrong diagnoses and yield systematic misdiagnoses that are difficult to detect.

To address this gap, we propose a training-free decision-point evidence sufficiency assessment and a three-tier sequential triage protocol for on-premise clinical AI agents (Fig. 1). When the local Qwen3.5-4B consulting doctor model finishes a consultation and is ready to submit a diagnosis, the system masks its reasoning process and proposed diagnosis and invokes the same local model without task-specific fine-tuning at temperature 0 to read the probability distribution across a five-level ordered scale—assessing whether the collected objective clinical evidence alone is sufficient to support a diagnosis—and compute an evidence sufficiency score without generating free text. Before deployment, the protocol locks a per-case threshold on a local calibration set according to a prespecified nominal auto-accept fraction and then evaluates, in ascending order of compute cost, a single-run tier, a five-run consistency tier, and a full-deferral tier that defers all cases to clinician review, certifying admission only when the one-sided Wilson lower confidence bound of the pass rate among accepted cases meets the clinical target line. We validated this framework across 1,605 primary-analysis cases spanning the primary acute-care benchmark MIRA-v2, the Clinical Decision-Making (CDM) acute abdominal-pain cohort also derived from MIMIC-IV, the cross-specialty open-consultation benchmark VivaBench, and an in-hospital liver malignancy primary cohort, further expanded into a combined enriched cohort. Together with consulting doctor model and reader model substitutions and medical-record information-ablation controls, we systematically compared evidence sufficiency assessment against token generation probability, direct 0–1 probability reporting, and five-run consistency.

Building on the general statistical calibration and held-out acceptance framework[26], this study advances on-premise clinical quality control and human–AI collaborative review in three respects. First, we establish an independent verification mechanism grounded in objective clinical evidence at the diagnostic submission point, preventing plausible-sounding reasoning narratives from distorting confidence judgments, thereby achieving a high pass rate among accepted cases in common acute conditions at single-run compute while eliminating tie clustering in the high-score range seen with self-reported diagnostic confidence. Second, distinct from cross-scale model cascades[30], we separate settings in which patient history and investigation results have already been compiled and multiple back-end inference calls are permitted from real-time outpatient settings with strict latency and local concurrency constraints: the former can invoke the five-run consistency tier with the same local model in the back end as a second tier, whereas the latter should avoid fivefold inference overhead and defer directly to clinician review if the single-run tier fails certification. Third, through cross-cohort empirical analysis, we distinguish diagnostic instability across closely related conditions—which the five-run consistency tier can correct—from missing objective evidence, enrichment of atypical pathological subtypes, or constrained cross-specialty differential capability that repeated sampling cannot remedy, providing hospitals with an actionable pathway for pre-deployment certification and timely transition to the full-deferral tier tailored to departmental case mix.

## Results

### Local judging validation and passive reliability signal baselines on the primary acute-care benchmark

We evaluated the triage protocol across 1,605 primary-analysis cases and confirmed high agreement between local automated judging and cloud re-judging on the primary acute-care benchmark MIRA-v2. The primary analysis comprised 551 acute-care cases from MIRA-v2[32–36], 599 cases from the same-source acute abdominal-pain cohort CDM[32,33,35,37], 274 cases from the cross-specialty open-consultation benchmark VivaBench[38], and 181 pathology-confirmed cases in the in-hospital liver malignancy primary cohort, which was further expanded to 514 cases to form the combined enriched cohort (Table 1). In multi-turn consultations across seven acute disease categories in MIRA-v2, the local Qwen3.5-4B consulting doctor model submitted a final diagnosis after collecting medical history, physical examination findings, laboratory tests, and imaging studies (Fig. 1a). In post hoc evaluation, the same local Qwen3.5-4B model served as the local 4B judge against the reference diagnosis, yielding a single-run consultation accuracy of 84.03%, which closely matched the 84.21% obtained by independent cloud re-judging with DeepSeek V4.1 Flash at a case-level agreement of 99.46%; local judging was therefore used uniformly in all subsequent primary analyses (Supplementary Fig. S11).

Comparing the two passive reliability signals on MIRA-v2 showed that five-run consistency achieved markedly higher triage discrimination than zero-overhead token generation probability, at the cost of five complete consultations per case. Specifically, token generation probability computes the geometric mean of predicted token probabilities across the submitted diagnosis sequence, whereas five-run consistency independently repeats the full consultation five times and outputs a single final diagnosis by consensus clustering without access to reference diagnoses, deferring non-consensus cases directly to clinician review. We defined the pass rate among accepted cases as the proportion of correctly diagnosed cases among auto-accepted cases, and the auto-accept fraction as the proportion of all cases spared clinician review. At a fixed top-50% auto-accept fraction, five-run consistency achieved a pass rate among accepted cases of 98.91%, with a 95% confidence interval (CI) of 96.85%–99.63%, which was markedly higher than the 89.82% achieved by token generation probability; at a pass-rate target of at least 98.0%, five-run consistency auto-accepted 61.34% of cases, compared with 0.00% for token generation probability (Fig. 2a).

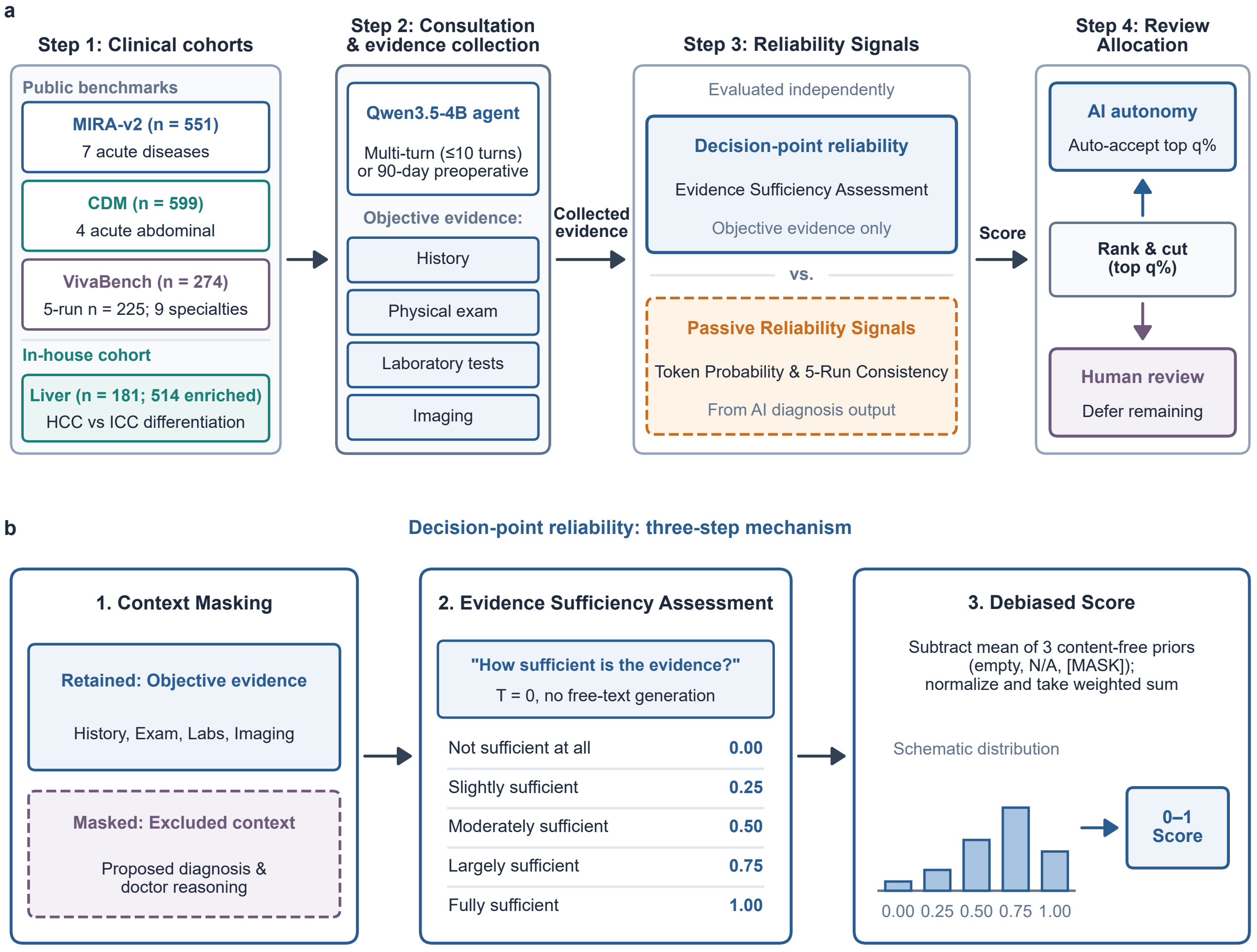

**Fig. 1 | System workflow and evidence sufficiency assessment mechanism. a**, Single-row four-step triage workflow. Step 1 includes three public benchmarks—MIRA-v2, CDM, and VivaBench—alongside the in-hospital liver malignancy primary cohort (*n* = 181) and combined enriched cohort (*n* = 514) differentiating HCC from ICC. Step 2 uses the Qwen3.5-4B doctor model to collect four categories of objective clinical evidence: history, physical examination, laboratory tests, and imaging studies. Step 3 compares two independent reliability assessment pathways in parallel: evidence sufficiency assessment in a blue box and passive signal comparators comprising token generation probability and five-run consistency in an orange box, outputting a single reliability score from the selected pathway. Step 4 ranks cases by score and sets a cutoff at a prespecified *q*% threshold, routing qualifying cases to the blue auto-accept branch and deferring the remainder to the gray-purple clinician review branch. **b**, Three-step mechanism of evidence sufficiency assessment. Step 1, context masking, retains the four categories of collected objective clinical evidence in a blue solid box while masking the proposed diagnosis and reasoning process in a gray-purple dashed box. Step 2, evidence sufficiency assessment, assigns equidistant scores from 0.00 to 1.00 across five ordered levels from "Not sufficient at all" to "Fully sufficient" and reads option probabilities without generating free text. Step 3, prior-debiased scoring, subtracts the mean of three content-free neutral priors—empty text, "N/A", and "[MASK]"—normalizes the distribution, and takes the weighted expectation to output a continuous score from 0 to 1; bars are schematic only. CDM, Clinical Decision-Making dataset; HCC, hepatocellular carcinoma; ICC, intrahepatic cholangiocarcinoma.

**Table 1 | Clinical cohorts, model configurations, full-cohort single-run accuracy, and judging agreement**

| Cohort | Sample size and task format | Consulting doctor model | Full-cohort single-run accuracy | Judging criterion and re-judging agreement |
|---|---|---|---|---|
| MIRA-v2 primary acute-care benchmark[32–36] | 551 cases (7 common acute conditions); multi-turn interactive consultation | Qwen3.5-4B | 84.03% (463/551) | Against reference diagnosis; cloud full-set agreement 99.46% (548/551) |
| Same-source acute abdominal-pain cohort CDM[32,33,35,37] | 599 cases (4 acute abdominal conditions); multi-turn interactive consultation | Qwen3.5-4B | 81.47% (488/599) | Against full ICD diagnosis; 100-case stratified cloud agreement 96.00% (96/100) |
| Cross-specialty open-consultation benchmark VivaBench[38] | 274-case three-doctor shared set (225-case five-run complete subset); multi-turn interactive consultation | Qwen3.5-4B / 27B doctor model / DeepSeek V4.1 Flash | 48.18% (132/274) / 66.79% (183/274) / 73.36% (201/274) | Benchmark-specific top-1 approximate criterion; doctor-stratified cloud agreement 85.77%–89.78% |
| In-hospital liver malignancy primary cohort | 181-case primary cohort (514-case combined enriched cohort); single-step static pre-operative record evaluation | Qwen3.5-4B / 27B doctor model* | 92.27% (167/181, complete version)† | Against pathology-confirmed HCC or ICC; first-line literal match rate 97.24%–97.79% |

*Note: Both the reader model and the primary local 4B judge across all cohorts were local Qwen3.5-4B, with the consulting doctor model sampling temperature set to 0.6 and the reader model temperature set to 0. Multi-turn interactive consultations were capped at 10 rounds and required at least 3 dialogue turns and 2 non-submission tool calls for laboratory tests or imaging studies; in CDM and the 225-case five-run complete subset of VivaBench, the five-run consistency tier was scored using a single final diagnosis by consensus. Independent cloud re-judging across the three public benchmarks used DeepSeek V4.1 Flash. All accuracy and agreement values in the table are point estimates for the corresponding full sets or re-judged subsets. *In the in-hospital liver cohort, the 27B doctor model was used solely for model-upgrade comparison on the 514-case complete version of usable pre-operative evidence, correctly diagnosing 431 of 514 cases for a full-cohort single-run accuracy of 83.85%. †"Complete version" denotes the complete version of usable pre-operative evidence; in the 181-case in-hospital liver primary cohort, 4B full-cohort single-run accuracy under the conclusion-masked and imaging-free versions was 86.19% (156 of 181 correct) and 79.01% (143 of 181 correct), respectively, whereas 4B complete-version full-cohort single-run accuracy on the 514-case combined enriched cohort was 80.35% (413 of 514 correct), as detailed in Supplementary Tables S3 and S5. CDM, Clinical Decision-Making dataset; HCC, hepatocellular carcinoma; ICC, intrahepatic cholangiocarcinoma; ICD, International Classification of Diseases.*

## Triage performance, threshold stability, and computational cost of evidence sufficiency assessment

Introducing evidence sufficiency assessment at the submission point of a single consultation matched the pass rate among accepted cases of five-run consistency without adding consultation rounds, while eliminating the triage-threshold uncertainty caused by high-score tie clustering under direct 0–1 probability reporting (Fig. 1b and Supplementary Table S2). At the fixed top-50% auto-accept fraction on MIRA-v2, evidence sufficiency assessment achieved a pass rate among accepted cases of 98.91% (95% CI 96.85%–99.63%) and intercepted 96.59% of all misdiagnoses into the clinician review group (Fig. 2d). By contrast, direct 0–1 probability reporting discretely clustered more than 60% of cases at the two highest values of 1.0 and 0.95, placing the top-50% threshold inside a large tie group, whereas the evidence sufficiency score yielded 551 distinct continuous values (Fig. 2b). Owing to these ties, under a 98.0% pass-rate target, the auto-accept fraction of direct 0–1 probability reporting fluctuated between 39.56% and 60.44% across 2,000 random tie-breaking permutations, and token generation probability never reached the target line, whereas evidence sufficiency assessment uniquely determined an auto-accept fraction of 59.53% (Fig. 2a).

Measured as complete consultation runs per correctly accepted case—the consultation compute required to autonomously accept one correct diagnosis—the single-run tier reduced consultation compute by approximately 80% relative to five-run consistency under both the clinical target line and high-safety targets. Under the 95.0% clinical target line, the single-run tier consumed only 1.34 consultation runs per correctly accepted case compared with 6.84 for five-run consistency; under the 98.0% high-safety target, consumption was 1.71 versus 8.30 runs, maintaining roughly an 80% compute reduction, with full cost breakdowns at higher target thresholds provided in the supplementary material (Fig. 2c and Supplementary Table S6). Moreover, the additional inference time for a single submission-point forward readout was less than 0.3% of one complete consultation, and its top-50% pass rate among accepted cases matched the endpoint of stepwise per-evidence readout conducted after every new piece of evidence throughout the consultation (Supplementary Fig. S8). Ablations on reader-model substitution, scale design, and joint stratification are provided in the supplementary material (Supplementary Figs. S1, S9, and S10).

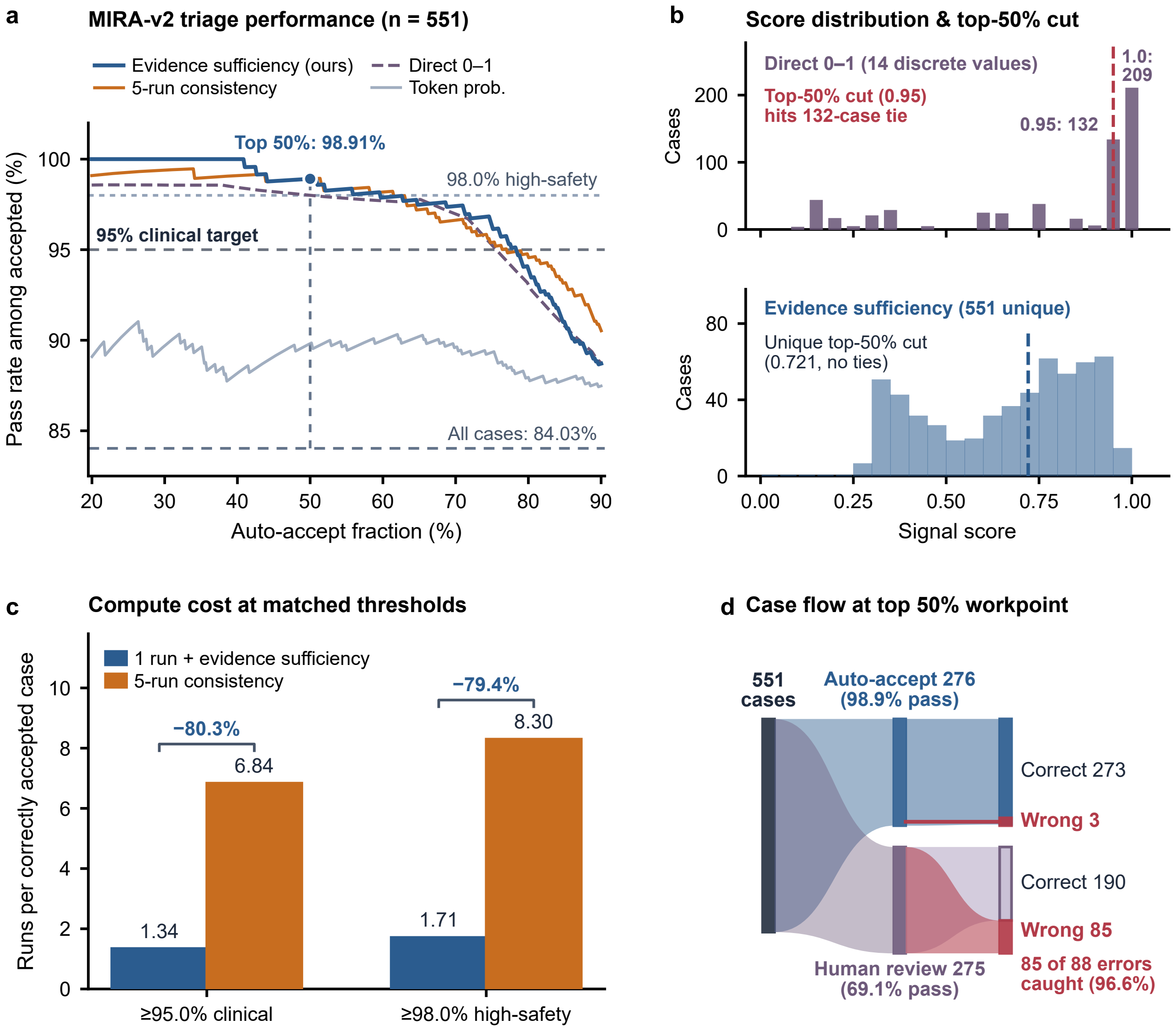

**Fig. 2 | Triage performance, score distributions, computational cost, and case flow on MIRA-v2. a**, Pass rate among accepted cases across 20%–90% auto-accept fractions (*n* = 551), with a zoomed *y*-axis. Solid blue and orange lines show evidence sufficiency assessment and five-run consistency defined by a single final diagnosis by consensus, respectively; the vertical dashed line and blue dot mark 98.91% at the top-50% fraction. The dashed purple line shows direct 0–1 probability reporting averaged over 2,000 random tie-breaking permutations, and the solid light-gray line shows token generation probability across 550 valid cases. Three horizontal lines denote the full-cohort single-run accuracy baseline, the 95% clinical target line, and the 98.0% high-safety line. **b**, Score distributions and top-50% thresholds with independent upper and lower *y*-axes. Upper purple bars show discrete clustering of direct 0–1 probability reporting, with the red dashed line marking the threshold inside a tied group; lower light-blue bars show continuous distinct values from the evidence sufficiency score, with the blue dashed line marking a unique threshold. **c**, Complete consultation runs per correctly accepted case under the 95.0% and 98.0% targets for the single-run tier (blue bars) versus five-run consistency (orange bars), showing ~80% compute savings. **d**, Sankey diagram at the top-50% fraction: the blue accepted group achieves a 98.9% pass rate, and the gray-purple clinician review group intercepts 96.6% of misdiagnoses. Full breakdowns appear in Supplementary Table S6.

## Validation on a same-source acute abdominal-pain cohort and single-run error composition

In the 599-case same-source Clinical Decision-Making (CDM) acute abdominal-pain cohort derived from MIMIC-IV—encompassing acute appendicitis, acute cholecystitis, colonic diverticulitis, and acute pancreatitis—overlapping clinical phenotypes brought the pass rate among accepted cases of the single-run tier close to the 95% clinical target line, whereas five-run consistency maintained robust separation (Fig. 3a). Here, the clinical target line denotes the minimum pass rate among accepted cases that auto-accepted cases must achieve, uniformly prespecified at 95% across all cohorts without dataset-specific adjustment. At the fixed top-50% auto-accept fraction, the CDM single-run tier achieved a pass rate among accepted cases of 96.67% (95% CI 93.97%–98.18%), exceeding the clinical target line by only 1.67 percentage points and falling below the 98.91% observed on MIRA-v2; at the same fraction, five-run consistency maintained a pass rate among accepted cases of 99.00% (95% CI 97.10%–99.66%). Although the single-run tier retained a marked computational advantage under the 95.0% target, its auto-accept fraction narrowed substantially relative to five-run consistency (Supplementary Table S7). Detailed static comparisons of the three single-run signals on CDM are reported in the supplementary material (Supplementary Table S6a and Supplementary Note S8.3).

The decline in pass rate among accepted cases for the single-run tier stemmed primarily from misdiagnoses of the three non-appendicitis abdominal conditions as acute abdominal diseases outside the four target entities, as well as diagnostic instability across closely related conditions when confronting atypical imaging findings. Across all 111 single-run misdiagnoses in CDM, 73.87% fell within the broader acute abdominal-pain spectrum outside the four target diseases rather than reflecting mutual confusion among the four targets (Fig. 3b). Acute appendicitis, which typically exhibits characteristic clinical features, achieved a single-run consultation accuracy of 95.82%, whereas the three more phenotypically overlapping conditions ranged from 70.15% to 73.46% (Supplementary Table S1). For example, in de-identified case 28641731 with a reference diagnosis of gallstones with acute cholecystitis, five independent runs obtained identical ultrasound findings showing gallstones with mild pericholecystic fluid, no gallbladder wall thickening, and a negative sonographic Murphy sign; the doctor model misdiagnosed biliary colic in the first run but correctly diagnosed acute cholecystitis in the remaining 4 runs. Because the collected objective evidence was sufficient, this case received a high evidence sufficiency score of 0.853 in the first run and entered the top-50% accepted group, showing that the single-run tier inspecting objective evidence alone cannot detect interpretive instability under atypical findings and requires the five-run consistency tier for correction.

### Cross-specialty open-consultation validation and consulting-doctor model substitution

On the cross-specialty open-consultation benchmark VivaBench spanning nine specialty groups, we evaluated the 274-case three-doctor shared set completed by all three consulting doctor models: local Qwen3.5-4B, a larger 27B doctor model, and cloud DeepSeek V4.1 Flash (Fig. 3c). Because this benchmark uses an approximate evaluation criterion that credits acceptable differential diagnoses, absolute accuracies are not directly comparable with those on the two acute-care cohorts; constrained by the cross-specialty differential capability of the local 4B model, both the single-run tier and the five-run consistency tier fell well below the 95% clinical target line and remained uncertified even after upgrading the consulting doctor model. At the fixed top-50% auto-accept fraction, the local 4B doctor model paired with evidence sufficiency assessment achieved a pass rate among accepted cases of only 55.47% (95% CI 47.12%–63.53%); in the 225-case five-run complete subset where all five independent runs finished normally with complete judging, five-run consistency also reached only 63.11% at the top-50% fraction. Holding the local 4B reader model fixed and replacing the consulting doctor model with the 27B doctor model or cloud DeepSeek V4.1 Flash raised the top-50% pass rate among accepted cases to 81.02% (95% CI 73.64%–86.71%) and 86.13% (95% CI 79.35%–90.94%), respectively, yet even tightening auto-acceptance to the top-20% fraction remained below the 95% clinical target line (Supplementary Note S8.2). Thus, the primary triage bottleneck lies in the evidence-gathering and differential-diagnosis capability of the consulting doctor model rather than the readout mechanism itself.

The failure of repeated consultations in cross-specialty open consultations arose because systematic differential-reasoning deficits in the consulting doctor model led independent runs to produce distinct wrong diagnoses. Among the 225-case five-run complete subset of the local 4B doctor model, 31 cases were misdiagnosed across all five runs, of which 23 yielded five mutually distinct wrong diagnoses—frequently substituting clinical syndromes such as heart failure or sepsis for definitive diagnoses, over-relying on nonspecific signs, or selecting the wrong entity within the same disease spectrum. For instance, in de-identified case pbm_26315723 with a reference diagnosis of central nervous system tuberculosis, 3 of the 5 independent runs detected cerebrospinal fluid tuberculosis PCR positivity yet misclassified the case as neurolyme disease, acute disseminated encephalomyelitis, and neurocysticercosis, respectively, while the other 2 runs failed to order the test and misdiagnosed stroke. Accordingly, when a consulting doctor model cannot establish the correct differential direction or translate confirmatory tests into the right diagnosis, neither evidence sufficiency assessment nor the five-run consistency tier can correct the reasoning bias, and the deployment protocol must halt auto-acceptance and defer all cases to clinician review; verification of the remaining 49 cases without five normal completions is detailed in the supplementary material (Supplementary Note S9).

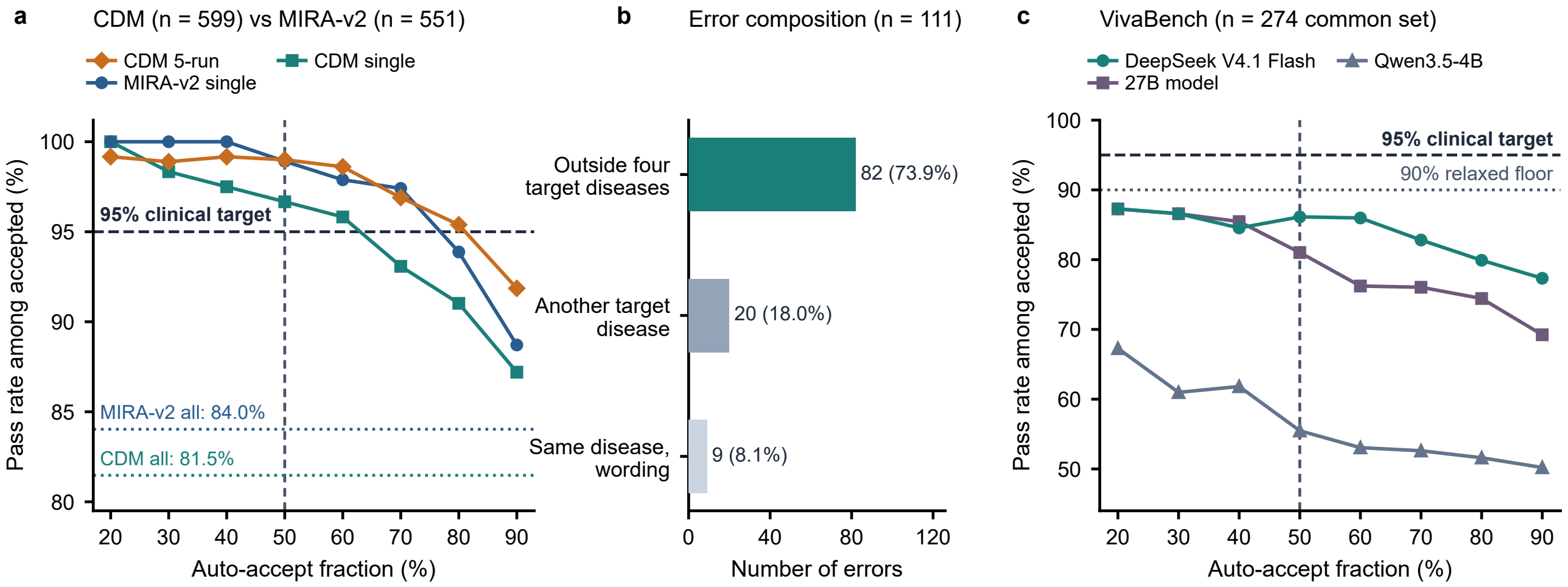

**Fig. 3 | Cross-cohort public benchmark triage performance and error composition. a**, Coverage–pass-rate curves comparing CDM (*n* = 599) and MIRA-v2 (*n* = 551). The solid orange line with diamonds shows CDM five-run consistency with a 99.00% pass rate among accepted cases at the top-50% fraction, the solid dark-blue line with dots shows the MIRA-v2 single-run tier at 98.91% at the top-50% fraction, and the solid teal line with squares shows the CDM single-run tier at 96.67% at the top-50% fraction. The horizontal dashed line marks the 95% clinical target line, and horizontal dotted lines mark full-cohort single-run accuracy baselines of 84.0% for MIRA-v2 and 81.5% for CDM. **b**, Clinical distribution of the 111 single-run misdiagnoses in CDM: teal horizontal bar shows 82 cases, or 73.9%, in the acute abdominal-pain spectrum outside the four target diseases; slate-blue horizontal bar shows 20 cases, or 18.0%, misclassified into another target disease; light-gray horizontal bar shows 9 cases, or 8.1%, reflecting wording or granularity differences within the same disease. **c**, Consulting doctor model substitution across the 274-case VivaBench three-doctor shared set with the local 4B reader model fixed. Solid teal line with dots, cloud DeepSeek V4.1 Flash; solid purple line with squares, 27B doctor model; solid slate-blue line with triangles, local Qwen3.5-4B. Horizontal dashed and dotted lines mark the 95% clinical target line and 90% relaxed admission floor, respectively. All curves show single-cohort point estimates without error bars; 95% confidence intervals at key fractions are given in the text and Supplementary Note S7.1. CDM, Clinical Decision-Making dataset.

## Three-tier sequential triage protocol and cross-cohort admission certification

When a nominal top-50% auto-accept fraction is fixed in advance on a local calibration set to lock a frozen per-case threshold and admission is certified against the 95% clinical target line using the one-sided 95% Wilson lower confidence bound of the pass rate among accepted cases, the three-tier sequential triage protocol achieves low-cost single-run tier auto-acceptance on MIRA-v2 and automatically switches to the five-run consistency tier on CDM to maintain target attainment (Fig. 4a, b). The nominal fraction must be prespecified and locked before deployment; searching backward on the calibration set for a threshold that maximizes coverage causes test-set attainment to drop substantially owing to finite-sample overfitting (Supplementary Note S1.1). In evaluation, each random split sequentially tests the single-run tier, five-run consistency tier, or full-deferral tier using the calibration set alone, with overall test-set attainment evaluated across all 200 splits, counting uncertified or full-deferral tier splits as non-attaining, and the calibration false-certification rate defined as the proportion of splits certified on the calibration set that fail to reach the 95% target line on the test set; mean consultation-equivalents also incorporate submission-point forward readout overhead under the conservative research-phase upper bound. Across 200 stratified random splits, the lower confidence bound of the MIRA-v2 single-run tier consistently exceeded the 95% clinical target line, yielding an all-split test attainment rate of 99.00%, a calibration false-certification rate of 0.00%, and a mean consumption of 1.30 consultation-equivalents per case; on CDM, because the single-run tier lower confidence bound frequently fell below the clinical target line, 82.5% of splits automatically routed to five-run consistency tier verification, restoring all-split test attainment to 96.00%, bounding the calibration false-certification rate at 3.50%, and consuming a mean of 4.54 consultation-equivalents per case (Table 2).

When clinical tasks exceed the model's differential capability boundary or when the clinical target line is tightened further, the sequential protocol halts high-risk auto-acceptance via the full-deferral tier. Because no configuration on VivaBench passed certification under the 95% clinical target line, we evaluated a relaxed admission floor of 90% in a sensitivity analysis; at the top-50% calibration auto-accept fraction, the local 4B model, the upgraded 27B doctor model, and cloud DeepSeek V4.1 Flash all had single-run tier pass rates among accepted cases and one-sided 95% Wilson lower confidence bounds below this 90% relaxed floor (Fig. 4b). All of these configurations had a 0.0% calibration certification pass rate and entered the full-deferral tier with zero false certifications (Table 2). Certification and information-ablation analyses for the in-hospital liver primary cohort are detailed in the next section. When raising the clinical target line for MIRA-v2 and CDM from 95% to 98% under the primary one-sided 95% confidence criterion ($z$ = 1.6449), the full-deferral rate of the sequential policy rose to 85.5% and 88.5%, respectively, indicating that at a 300-case calibration size, 95% represents the upper target line that can be robustly certified on both acute cohorts (Supplementary Note S1.2).

**a Three-tier sequential triage protocol**

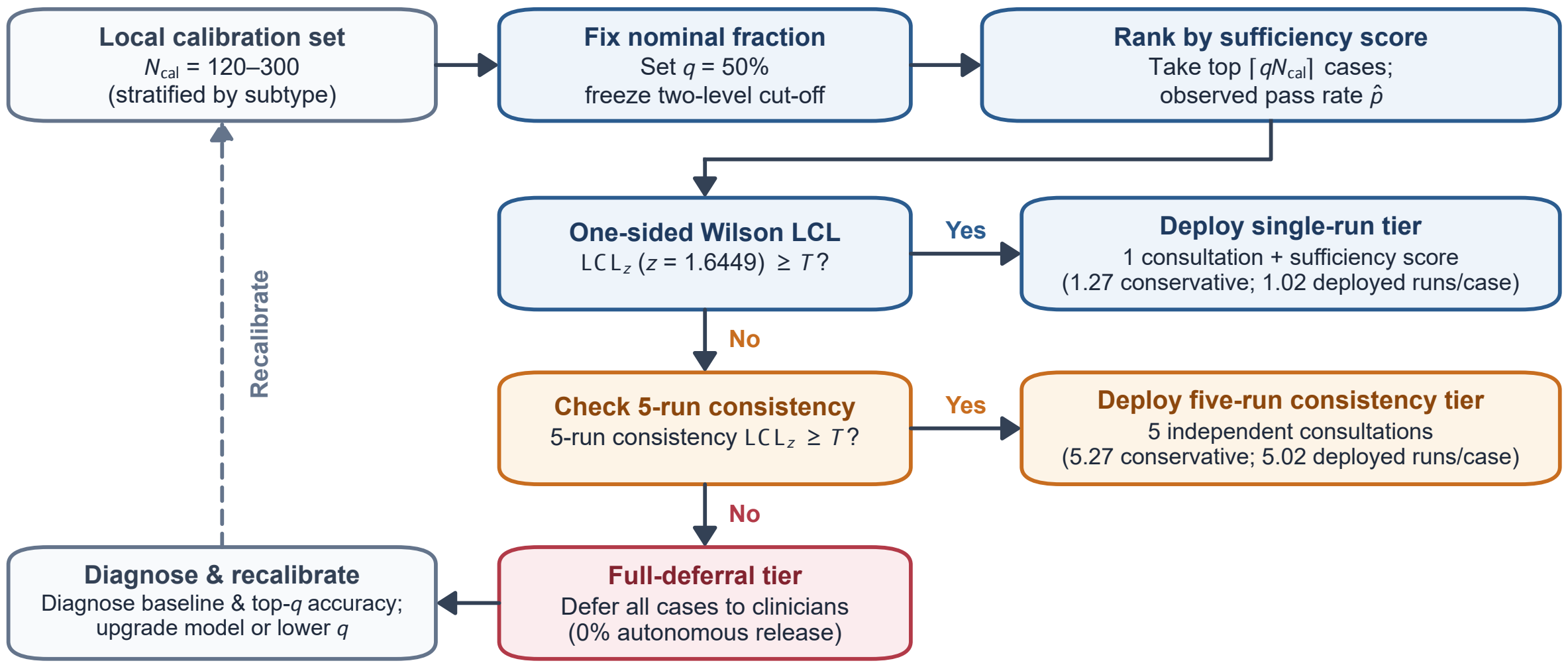

**b Calibration top-50% point estimates and one-sided 95% Wilson lower bounds**

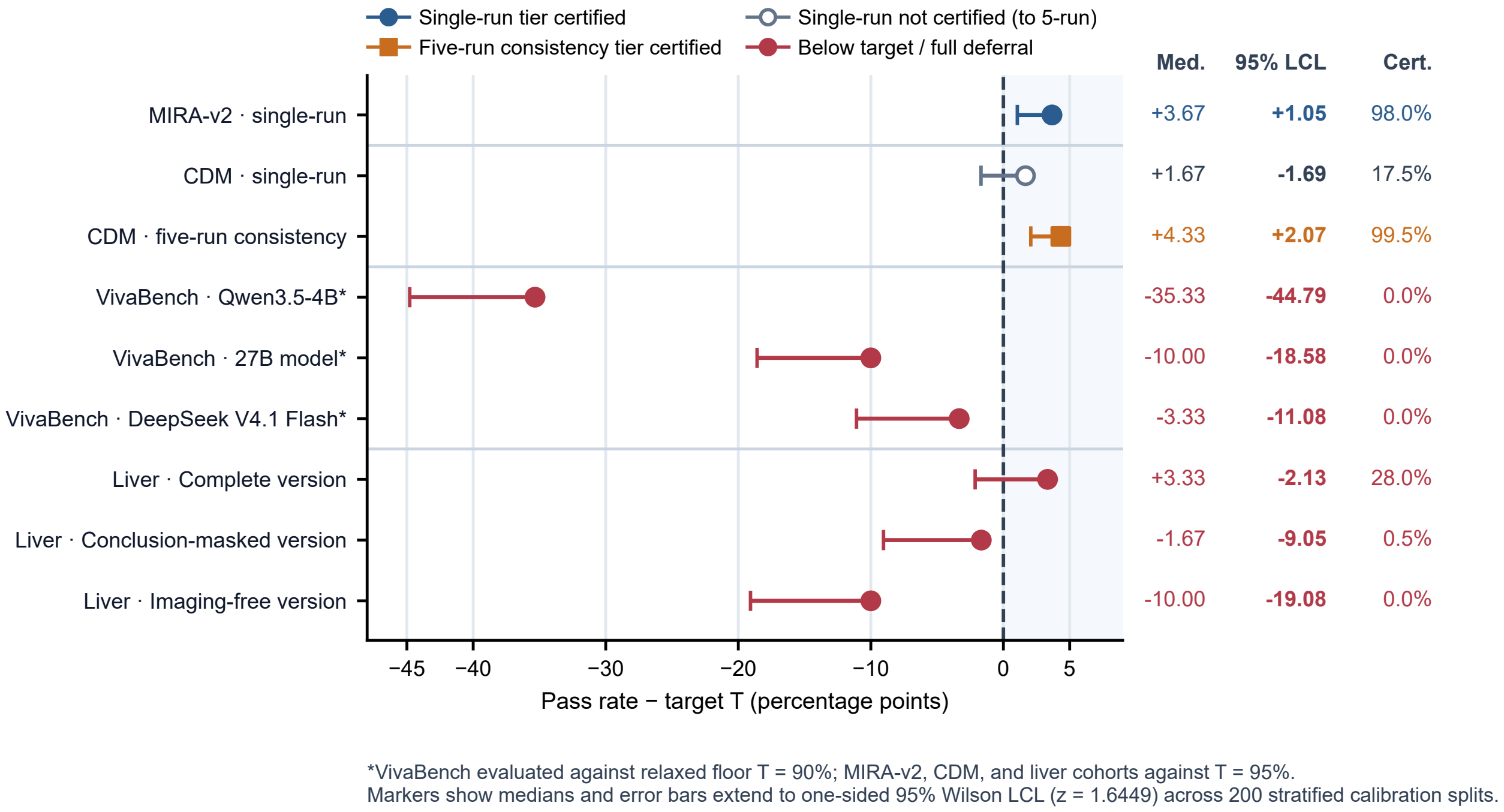

**Fig. 4 | Workflow of the three-tier sequential triage protocol and cross-cohort calibration attainment differences. a**, Decision flow of the three-tier sequential triage protocol. On a subtype-stratified local calibration set of 120–300 cases, a nominal auto-accept fraction $q$ of 50% is prespecified, cases are ranked by the evidence sufficiency score, and a frozen per-case threshold is locked. Using attainment of the clinical target line $T$ by the one-sided 95% Wilson lower confidence bound of the pass rate among accepted cases as the criterion, the protocol sequentially evaluates the single-run tier in blue boxes and the five-run consistency tier in orange boxes; if neither meets $T$, the full-deferral tier in a red box defers all cases to clinician review and initiates root-cause investigation and recalibration in a gray box. **b**, Median percentage-point difference between the pass rate among accepted cases and the clinical target line $T$ at the top-50% auto-accept fraction across 200 stratified random calibration splits, with $T$ = 95% for MIRA-v2, CDM, and the in-hospital liver primary cohort, and a relaxed admission floor of $T$ = 90% for cross-specialty VivaBench. Points show median differences: filled blue circles denote certified single-run tiers, open gray-blue circles uncertified single-run tiers, filled orange squares certified five-run consistency tiers, and filled dark-red circles failure to attain the target or entry into the full-deferral tier. Leftward horizontal error bars denote the median one-sided 95% Wilson lower confidence bound, and the three right-hand columns list the median difference, lower confidence bound, and calibration certification pass rate. CDM, Clinical Decision-Making dataset; LCL, one-sided 95% Wilson lower confidence bound.

**Table 2 | Results of 200 sequential deployment simulation splits across four clinical cohorts at the nominal top-50% auto-accept fraction under the primary one-sided 95% confidence criterion**

| Cohort and calibration configuration | Candidate single-run tier certification pass rate | All-split test attainment rate (sequential policy) | Calibration false-certification rate | Full-deferral rate | Mean consultation-equivalents* |
|---|---|---|---|---|---|
| MIRA-v2 (*n* = 551; calibration 300 cases; *T* = 95.0%) | 98.0% (196/200) | 99.00% (198/200) | 0.00% (0/200) | 1.00% (2/200) | 1.30 |
| CDM (*n* = 599; calibration 300 cases; *T* = 95.0%)† | 17.5% (35/200) | 96.00% (192/200) | 3.50% (7/200) | 0.50% (1/200) | 4.54 |
| VivaBench (*n* = 274; calibration 150 cases; floor *T* = 90.0%)‡ | 0.0% (0/200) | 0.00% (0/200) | 0.00% (0/200) | 100.00% (200/200) | 0.00 |
| In-hospital liver primary cohort, complete version (*n* = 181; calibration 120 cases; *T* = 95.0%)§ | 28.0% (56/200) | 28.00% (56/200) | 0.00% (0/200) | 72.00% (144/200) | 0.36 |

*Note: Proportions in all columns use all 200 stratified random splits as the denominator, and mean consultation-equivalents represent the mean across all 200 splits; the all-split test attainment rate, calibration false-certification rate, and full-deferral rate sum to 100%. Admission certification used the primary one-sided 95% Wilson lower confidence bound criterion (z = 1.6449); results under the lenient comparator (z = 1.0, corresponding to approximately 84% one-sided confidence) and five mutually exclusive split outcomes are detailed in Supplementary Note S1.3. *Under conservative research-phase accounting, the single-run tier, five-run consistency tier, and full-deferral tier are counted as 1.27, 5.27, and 0 consultation-equivalents, respectively; under actual deployment accounting with a single submission-point forward readout (1.02 and 5.02 consultation-equivalents), mean consumption was 1.05 for MIRA-v2 and 4.29 for CDM. †The CDM five-run consistency tier was scored using a single final diagnosis by consensus, yielding an independent candidate certification pass rate of 99.5%; within the sequential policy, of the 165 splits that failed single-run tier calibration certification and entered five-run consistency tier verification, 164 passed calibration certification and 1 entered the full-deferral tier. ‡On VivaBench, the single-run tiers for local Qwen3.5-4B, the 27B doctor model, and cloud DeepSeek V4.1 Flash, as well as the five-run consistency tier on the 225-case five-run complete subset, all entered the full-deferral tier. §"Complete version" denotes the complete version of usable pre-operative evidence; full comparisons among the complete, conclusion-masked, and imaging-free versions of the in-hospital liver cohort are provided in Fig. 5 and Supplementary Table S4. Under the primary criterion, the full-deferral rates for the conclusion-masked and imaging-free versions were 98.50% and 100.00%, respectively, with calibration false-certification rates of 1.50% and 0.00% (Supplementary Note S1.3). CDM, Clinical Decision-Making dataset;* n, *total cohort sample size;* T, *clinical target line (with a relaxed admission floor of 90.0% set for VivaBench).*

## Sequential triage and full-deferral boundaries in the in-hospital liver malignancy primary cohort

Applying the three-tier sequential triage protocol to the 181-case pathology-confirmed in-hospital liver malignancy primary cohort showed that the high pass rate among accepted cases of the single-run tier depended heavily on explicit qualitative diagnostic conclusions from radiologists, and that its pathological spectrum and single-step task format differ from real-world outpatient or screening populations. To test whether the protocol halts auto-acceptance in time under three real-world in-hospital risks, we extracted all pre-operative clinical evidence within a 90-day pre-procedure window without truncation and constructed, in addition to the complete version of usable pre-operative evidence, a conclusion-masked version that stripped radiologists' qualitative conclusions while retaining objective imaging findings, an imaging-free version retaining only demographics and laboratory tests to simulate missing key imaging modalities, and a 514-case combined enriched cohort with elevated intrahepatic cholangiocarcinoma prevalence to test subtype shift. For each input version under the single-run tier, the local Qwen3.5-4B reader model performed evidence sufficiency assessment solely on the corresponding pre-operative objective clinical evidence while masking the doctor model's diagnosis and rationale, ranking cases and auto-accepting the top 50%. At this fraction, the 181-case complete version achieved a pass rate among accepted cases of 98.90% (95% CI 94.04%–99.81%; 2,000-permutation one-sided test against random triage, *P* = 0.0005), yet all auto-accepted cases were hepatocellular carcinoma, and full-cohort single-run accuracy matched a simple keyword-only rule baseline searching for cancer terms in imaging reports (Supplementary Table S3). By contrast, owing to increased hepatocellular carcinoma misclassifications and misdiagnosed intrahepatic cholangiocarcinoma cases entering the accepted group, the top-50% pass rate among accepted cases of the single-run tier fell to 94.51% on the conclusion-masked version and 84.62% on the imaging-free version—both below the 95% clinical target line—with exact permutation *P* values for all input versions detailed in the supplementary material (Fig. 5a and Supplementary Table S3).

Unlike the CDM acute abdominal-pain cohort—where the five-run consistency tier could correct random interpretive instability—when misdiagnoses stem from missing key imaging conclusions or enrichment of challenging pathological subtypes, five-run consistency merely reproduces the same objective-evidence bias and must be intercepted by the full-deferral tier. In the conclusion-masked and imaging-free versions, multiple hepatocellular carcinoma and intrahepatic cholangiocarcinoma cases ranked into the top-50% group by evidence sufficiency assessment remained consistently misdiagnosed across five-run consistency review, preventing the five-run consistency tier from eliminating stable errors; moreover, applying the lenient comparator ($z$ = 1.0), corresponding to approximately 84% one-sided confidence, to the conclusion-masked version yielded a calibration false-certification rate of 22.50%, with none of the certified splits attaining the target on the test set (Fig. 5a and Supplementary Note S1.3). On the 514-case combined enriched cohort, even with the complete version retaining unmasked imaging conclusions, the top-50% pass rate among accepted cases of the single-run tier declined to 88.33%, and stratified resampling curves based on the complete version and evidence sufficiency assessment confirmed that pass rates among accepted cases decreased continuously as the share of intrahepatic cholangiocarcinoma increased, failing to reach the 95% clinical target line even when the single-step consulting doctor model was upgraded to the 27B doctor model (Fig. 5b and Supplementary Table S5). Confronting such evidence gaps and subtype-distribution shifts that repeated sampling cannot remedy, admission certification of the single-run tier on the complete-version resampled cohorts using the primary one-sided 95% confidence criterion ($z$ = 1.6449) rapidly tightened certification and routed splits to the full-deferral tier as intrahepatic cholangiocarcinoma prevalence rose (Fig. 5c). Across all evidence-ablation and subtype-enrichment scenarios, this primary criterion triggered the full-deferral tier in 98.50%–100.00% of splits and bounded the calibration false-certification rate at 0.00%–1.50% across all in-hospital evaluation settings, including the complete version (Supplementary Table S4).

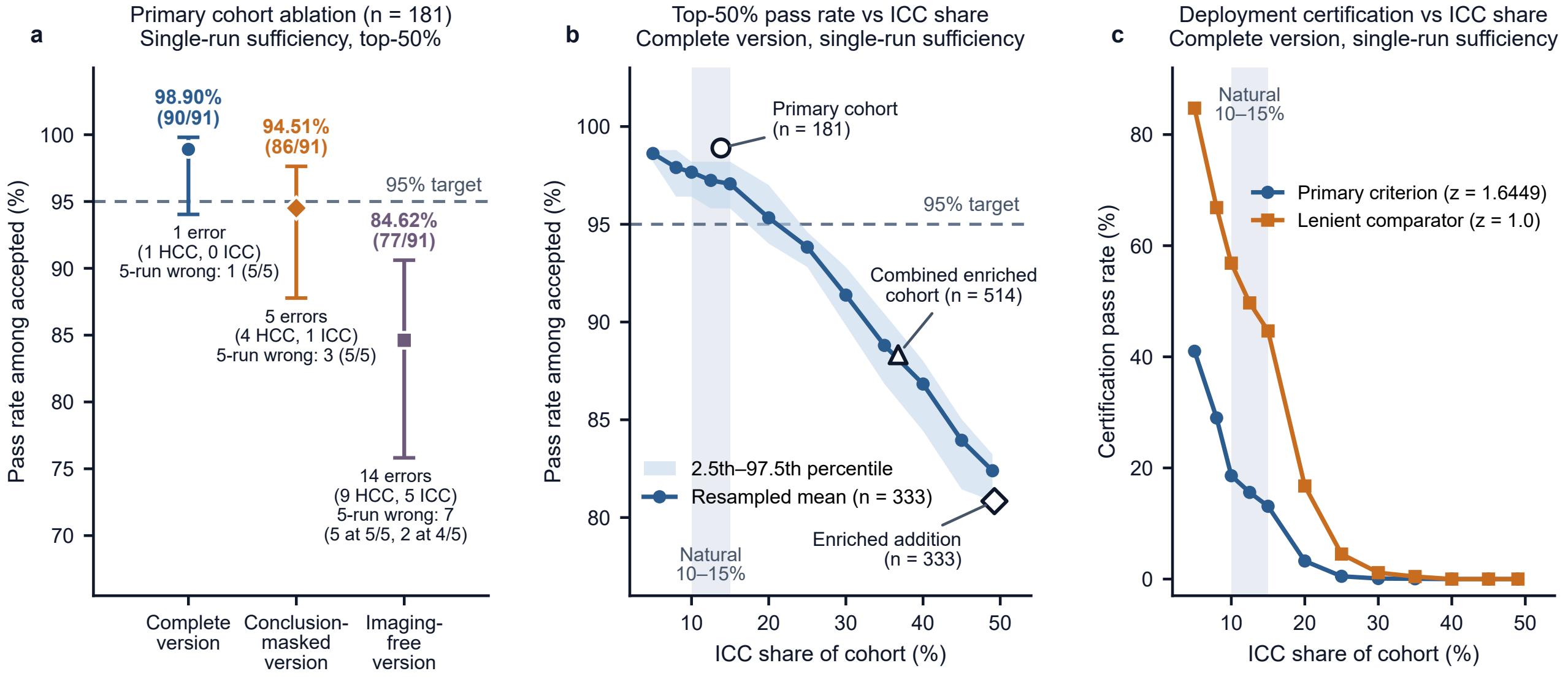

**Fig. 5 | Information-ablation and subtype-enrichment full-deferral boundaries in the in-hospital liver cohort.** All panels use the single-run tier with local Qwen3.5-4B evidence sufficiency assessment, which evaluates only pre-operative objective clinical evidence while masking the doctor model's diagnosis and rationale and auto-accepts the top-ranked 50% of cases. **a**, Pass rate among accepted cases at the top-50% auto-accept fraction across three information-ablation inputs in the 181-case pathology-confirmed in-hospital liver malignancy primary cohort: blue circle, complete version of usable pre-operative evidence at 98.90%; orange diamond, conclusion-masked version at 94.51%; gray-purple square, imaging-free version at 84.62%. Vertical error bars show two-sided 95% Wilson confidence intervals, and the horizontal dashed line marks the 95% clinical target line. In-plot annotations indicate total single-run errors in the accepted group, the breakdown into hepatocellular carcinoma and intrahepatic cholangiocarcinoma misdiagnoses, and the number and consistency of cases remaining majority-incorrect across five independent runs. **b**, Pass rate among accepted cases at the top-50% fraction across 12 target ICC-share gradients from 5.0% to 49.0%, constructed by stratified sampling without replacement of 10 simulated cohorts of 333 cases each from the complete version with imaging conclusions retained of the 514-case combined enriched cohort. The blue solid line and light-blue shading show the mean and 2.5%–97.5% quantile interval, the vertical light-gray band marks the literature-reported 10%–15% natural ICC prevalence range among primary liver cancers[39,40], and the open circle, open triangle, and open diamond denote observed complete-version points for the 181-case primary cohort, 514-case combined enriched cohort, and 333-case newly added enrichment set. **c**, Mean calibration certification pass rate across 200 random splits for the 10 complete-version simulated cohorts at each ICC share: blue circles, primary one-sided 95% confidence criterion with $z$ = 1.6449; orange squares, lenient comparator with $z$ = 1.0. HCC, hepatocellular carcinoma; ICC, intrahepatic cholangiocarcinoma.

## Discussion

We introduced a training-free evidence sufficiency assessment combined with fixed-fraction local calibration and one-sided 95% Wilson lower confidence bound certification, keeping the calibration false-certification rate within 0.00%–3.50% across six deployment evaluation scenarios spanning the three public benchmarks MIRA-v2, CDM, and VivaBench as well as the complete, conclusion-masked, and imaging-free versions of the in-hospital liver primary cohort. Depending on cohort difficulty and evidence quality, this sequential protocol automatically differentiated into three admission behaviors: on the acute-care benchmark MIRA-v2, where baseline differential capability is sufficient, the single-run tier reduced complete consultation runs per correctly accepted case by approximately 80%; on the phenotypically overlapping CDM acute abdominal-pain cohort, the protocol automatically routed to the five-run consistency tier to correct diagnostic instability under atypical findings; and on cross-specialty VivaBench as well as in-hospital liver cohort scenarios with missing imaging conclusions or enrichment of challenging pathological subtypes, the protocol promptly triggered the full-deferral tier to defer all cases to clinician review. Together, reliability in clinical AI agents shifts from a static output property of a model to a clinical admission boundary jointly defined by cohort characteristics, a nominal auto-accept fraction, and a statistical confidence threshold that can be estimated and verified prior to deployment.

The effectiveness of evidence sufficiency assessment stems from inspecting only collected objective clinical evidence while completely isolating the consulting doctor model's reasoning process and proposed diagnosis, thereby blocking plausible-sounding reasoning narratives from distorting sufficiency judgments (Supplementary Fig. S9c). Cross-model controls showed that holding the local Qwen3.5-4B doctor model's consultation trajectories fixed and replacing only the reader model with Qwen3.5-27B or Gemma 4 E4B did not yield a marked improvement in ranking discrimination (Supplementary Fig. S1). Notably, Qwen3.5-27B exhibited score compression toward middle levels and a narrowed score span, indicating that score dispersion should be verified before deploying a larger reader model (Supplementary Note S8.1). Conversely, upgrading the consulting doctor model on VivaBench to the 27B doctor model or cloud DeepSeek V4.1 Flash markedly improved the pass rate among accepted cases (Fig. 3c). These findings indicate that triage performance is bounded primarily by the consulting doctor model's differential-diagnosis capability and the quality of objective clinical evidence, rather than by scaling up the reader model alone. From a hospital deployment perspective, the approach is training-free, adds no consultation rounds, reduces consultation compute by approximately 80% relative to five-run consistency, and yields deterministic, auditable, and reproducible probability readouts at temperature 0 whenever the collected objective evidence is fixed. Collectively, while the consulting doctor model and objective evidence define the upper bound on auto-acceptance, objective-evidence isolation enables the single-run tier to achieve both high discrimination and auditable determinism at no additional consultation cost.

Our primary contribution lies in coupling objective-evidence-isolated assessment, continuous probability-distribution scoring, and three-tier local statistical admission into a mutually reinforcing on-premise triage architecture. First, reading a prior-debiased probability distribution over a five-level ordered scale at the diagnostic submission point and taking its weighted expectation expands high-score tie clusters seen in direct numerical probability generation into distinct, continuous case-level evidence sufficiency scores, while avoiding both the coarseness of binary assessment and the dilution of finer-grained scales, thereby providing stable thresholds at single-run compute. Second, for deployment certification, prespecifying the nominal auto-accept fraction and locking a frozen per-case threshold before launch prevents opportunistic overfitting caused by searching backward for maximum coverage on the local calibration set; gating admission with the one-sided 95% Wilson lower confidence bound ensures that limited calibration sample sizes or minor contamination in the accepted group naturally depress the lower bound and trigger tier escalation or deferral, enabling a single on-premise small model to transition seamlessly across the single-run tier, five-run consistency tier, and full-deferral tier without model replacement. Finally, integrating cross-cohort validation into the same certification framework allows hospitals to separate trajectory-level interpretive fluctuations absorbable by repeated sampling from capability or evidence gaps that mandate halting auto-acceptance prior to launch. Thus, continuous submission-point scoring, frozen-threshold calibration, and one-sided confidence gating act together to reconcile low-compute autonomy with rigorous admission control in a single on-premise model.

Empirical comparisons across the three cohort types clearly delineate three distinct failure boundaries: random fluctuation in diagnostic interpretation, insufficient cross-specialty differential capability, and missing objective evidence or enrichment of challenging pathological subtypes. First, although CDM shares its data source with MIRA-v2, the three non-appendicitis abdominal conditions were frequently misdiagnosed as acute abdominal entities outside the four target diseases. Faced with atypical imaging signs pointing simultaneously toward closely related diagnoses, evidence sufficiency assessment without seeing the proposed diagnosis can verify only that objective evidence is sufficient, without detecting interpretive instability across neighboring entities; this brings the single-run tier pass rate among accepted cases close to the clinical target line and requires switching to the five-run consistency tier to stabilize diagnoses by consensus. Second, in VivaBench cross-specialty open consultations, the lightweight doctor model failed to establish the correct differential direction or convert confirmatory tests into accurate final diagnoses, causing repeated runs to diverge across distinct wrong hypotheses. Because upgrading the consulting doctor model still fell short of the admission floor, only the full-deferral tier could prevent unsafe auto-acceptance. Consequently, random interpretive variation among closely related diagnoses can be absorbed through the five-run consistency tier with the same model, whereas tasks exceeding a local model's cross-specialty differential boundary require entering the full-deferral tier altogether.

The third failure boundary arises from gaps in objective evidence itself or shifts in pathological subtype composition, as demonstrated in the in-hospital liver cohort. Masking radiologists' qualitative conclusions or enriching atypical intrahepatic cholangiocarcinoma cases both increased the intrinsic misdiagnosis rate within the accepted group. In the combined enriched cohort, when pre-operative non-invasive clinical evidence lacked key information or favored the alternative subtype, even upgrading the consulting doctor model to the 27B doctor model could not eliminate misdiagnoses in the accepted group (Supplementary Table S5 and Supplementary Note S11.2). Moreover, repeated runs merely replicated the same objective-evidence bias and produced identical wrong diagnoses, rendering the five-run consistency tier ineffective against systematic errors. Stratified resampling curves showed that as the intrahepatic cholangiocarcinoma share rose from the literature-reported 10%–15% natural prevalence range among primary liver cancers[39,40] to 20% or higher, the pass rate among accepted cases approached or dropped below the 95% clinical target line (Fig. 5b). Under these conditions, only the strict one-sided 95% Wilson lower confidence bound promptly halted auto-acceptance and entered the full-deferral tier, preventing systematic false certifications that occur under the lenient comparator (Fig. 5c and Supplementary Note S1.3). When misdiagnoses originate from missing objective evidence or atypical subtype enrichment, repeated sampling cannot compensate for inherent evidence deficits, making strict lower-bound certification essential for timely referral to clinician review.

In terms of clinical deployment and human-factors design, this approach should be prioritized for clinical workflows with well-defined diagnostic pathways, distinct clinical phenotypes, low risk of confusion among closely related entities, and complete key laboratory and imaging evidence—such as automated triage of common acute presentations and quality control of standardized specialty records—while settings with highly overlapping phenotypes, complex cross-specialty consultations, or missing diagnostic conclusions should route to the five-run consistency tier or enter the full-deferral tier to defer all cases to clinician review. Deployment must strictly distinguish two operational modes: the five-run consistency tier is suited only to workflows in which patient history and investigation results are already compiled and multiple back-end inference calls are permissible, such as back-end review after emergency laboratory and imaging workups or pre-operative inpatient record auditing; by contrast, in real-time outpatient consultations under tight response-latency and local concurrency limits, systems should not incur fivefold inference overhead and should defer directly to clinician review whenever the single-run tier fails certification. Furthermore, because nearly all misdiagnoses in MIRA-v2 at the top-50% fraction are routed to the clinician review group, presenting error-prone model diagnoses directly to busy clinicians risks inducing automation bias and preconceived diagnostic bias[41]. Clinician review interfaces should therefore preserve objective-evidence isolation by hiding the model's proposed diagnosis and rationale by default, displaying only structured objective evidence and low-reliability alerts, while hospital administrators can calibrate the nominal auto-accept fraction, clinical target line, and one-sided confidence level using decision curve analysis and asymmetric clinical harm weights[42]. Matching triage tiers to departmental evidence completeness and compute constraints, alongside hiding proposed diagnoses in clinician review interfaces, is central to translating this protocol into a safe clinical workflow.

Interpretation of these findings remains subject to methodological and statistical sampling boundaries. First, deployment certification simulations relied on retrospective random splits without replacement from finite samples, meaning splits are not fully independent; in small accepted groups, pass rates among accepted cases exhibit discrete step changes constrained by integer case counts, and sampling without replacement introduces negative correlation between calibration and test sets. Because retrospective simulations reflect only pre-launch statistical admission stability rather than prospective interventional outcomes, deployed systems require ongoing post-launch monitoring of error rates in the accepted group and subtype shifts, with timely recalibration or reversion to the full-deferral tier. Second, determinism at temperature 0 guarantees reproducibility only for a fixed consultation trajectory and does not remove sampling stochasticity across multi-turn consultation generation; absolute score thresholds, assessment prompt wording, and sampling temperatures cannot be transferred directly across institutions without local fixed-fraction calibration on site-specific samples (Supplementary Fig. S5). Additionally, the five-run consistency tier relies on sentence-embedding clustering to produce a single final diagnosis; although robust across similarity thresholds of 0.70–0.90 and complete-linkage clustering, it remains a heuristic merging rule (Supplementary Note S1.4). Consequently, retrospective pre-deployment certification cannot substitute for ongoing post-launch error monitoring and institutional recalibration, nor can single-site thresholds be transferred directly across hospitals.

Several model, cohort, and cost-accounting limitations should also be noted. Our evaluation centered on a local 4B consulting doctor model with patient simulators synchronized whenever consulting doctor models were substituted, and further validation across additional model architectures is still required; within-model two-stage cascading showed negligible gain, whereas cross-scale cascading and submission-point-only resampling remain to be systematically evaluated (Supplementary Note S8.3). In VivaBench, 49 cases lacking five complete normal completions were excluded from the five-run complete subset (Supplementary Note S9), and MIRA-v2 and CDM share partial source-record overlap, although removing overlapping hospitalizations yielded consistent conclusions (Supplementary Note S10). The in-hospital liver cohort was derived from a single center, lacked independent blinded physician re-review of reference labels, included multifocal lesions, excluded benign space-occupying lesions and combined hepatocellular–cholangiocarcinoma, and represents a retrospective pre-operative pathology-confirmed differential cohort rather than an outpatient or primary-care screening population. Finally, compute accounting captured only local model inference overhead and did not quantify clinician review time. Taken together, these model-architecture, retrospective-cohort, and cost-accounting boundaries highlight validation across additional model architectures, prospective outpatient clinical evaluation, and incorporation of clinician review time as key priorities for future clinical translation.

In summary, coupling a training-free, objective-evidence-isolated evidence sufficiency assessment with fixed-fraction local calibration, one-sided 95% Wilson lower confidence bound certification across the single-run tier, five-run consistency tier, and full-deferral tier, and evidence-isolated clinician review interfaces offers on-premise clinical AI agents an auditable, compute-bounded deployment pathway that prioritizes auto-acceptance where safe and promptly defers to clinician review where risks emerge.

## Methods

### Study design and clinical cohorts

We adopted a stepwise expansion design to construct clinical cohorts across four sequential validation levels: primary acute-care benchmark reproduction, same-source acute abdominal-pain validation, cross-specialty open-consultation validation, and in-hospital real-world retrospective information-ablation and subtype-enrichment validation; the study followed the TRIPOD+AI reporting guidelines[43] and did not prospectively recruit participants (Table 1). The primary acute-care benchmark MIRA-v2 was reconstructed from MIMIC-IV v2.2, MIMIC-IV-ED v2.2, and MIMIC-IV-Note v2.2 containing discharge summaries and imaging reports[32–35] using the publicly released codebase of the MIRA framework[36]. Following prior on-premise clinical AI agent evaluation criteria[11], we excluded 23 pancreatic cancer cases whose diagnosis depended heavily on cross-sectional imaging staging from the original 574 cases, yielding 551 cases spanning seven common acute conditions. In addition, a 524-case paired subset was partitioned from early development of MIRA-v2 solely for scale-design and reader-model substitution ablations and was excluded from primary outcome evaluations (Supplementary Figs. S1 and S9).

The same-source acute abdominal-pain cohort CDM was reconstructed from MIMIC-IV v2.2 and MIMIC-IV-Note v2.2[32,33,35] using official scripts provided by the public resource[37], yielding 2,389 cases; the 11-case difference relative to the 2,400 cases reported in the public resource reflected natural-language-processing dependency version differences, corresponding to a 99.54% case retention rate. Using a fixed random seed, we drew a stratified random sample of 600 cases proportional to disease composition and excluded 1 case lacking complete five-run consultation records, resulting in a 599-case analysis set comprising 239 cases of acute appendicitis, 162 of acute cholecystitis, 64 of colonic diverticulitis, and 134 of acute pancreatitis. A sensitivity analysis excluding 88 hospitalization records overlapping with MIRA-v2 is provided in the supplementary material (Supplementary Note S10).

The cross-specialty open-consultation benchmark VivaBench was derived from the publicly reviewed 990-case benchmark spanning nine specialty groups[38]. Using specialty-stratified sampling, we selected 274 cases completed by all three consulting doctor models—local Qwen3.5-4B, the 27B doctor model, and cloud DeepSeek V4.1 Flash—to form the three-doctor shared set for evaluating the single-run tier and cross-doctor comparisons. Within this shared set, 225 cases yielded complete termination markers and full judging results across all five independent runs, forming the five-run complete subset for evaluating the five-run consistency tier; baseline verification of the remaining 49 cases without normal five-run completions is detailed in the supplementary material (Supplementary Note S9).

The in-hospital liver malignancy primary cohort originated from a pre-established 200-case independent validation candidate library; after excluding 19 cases used for preliminary system debugging from the same library, we included 181 cases confirmed by surgical or biopsy pathology, comprising 156 cases of hepatocellular carcinoma and 25 of intrahepatic cholangiocarcinoma, 12 of which described multifocal lesions on final pathology. The study methods and analysis status for the benchmarks and 181-case primary cohort were retrospectively registered on the Open Science Framework on 12 November 2026 under DOI 10.17605/OSF.IO/S5DMH; this record documents completed analyses rather than prospective preregistration. Following registration, using identical inclusion and de-identification procedures, we additionally enrolled 333 pathology-confirmed cases, comprising 169 hepatocellular carcinoma and 164 intrahepatic cholangiocarcinoma, and merged them with the in-hospital liver primary cohort to form a 514-case combined enriched cohort. For every case, time zero, $T_0$, was defined as 23:59:59 on the day preceding the target tissue-sampling or antitumor procedure, and only in-hospital pre-operative records reported within 90 days before $T_0$, corresponding to −89 to 0 days relative to $T_0$, were included, strictly excluding any clinical data or pathology results from the procedure day onward.

Record texts supplied to the models retained only the de-identified study ID, age, sex, imaging reports, and laboratory test records. Imaging reports preserved examination type, body site, imaging findings, imaging conclusions, and original report text, retaining only conclusions when original text and conclusions contained redundant overlap, whereas laboratory test records preserved test panel, analyte name, numerical or qualitative value, unit, reference range, and relative report day, with missing values uniformly labeled "not obtained". During generation of the final model inputs, target-lesion curation citations originally embedded in task prompts for 135 records were removed in full while preserving the original clinical descriptions inside the corresponding imaging reports. All clinical evidence items were ordered chronologically by relative report day and evidence index on separate lines without truncation; across the 181-case in-hospital liver primary cohort, there were 873 imaging reports and 28,304 laboratory test records, with a median formatted input length of 12,558 characters and a maximum length of 32,114 characters.

From these standardized record texts, we constructed three parallel retrospective information-ablation versions under identical $T_0$, evidence indices, and relative report days. The first was the complete version of usable pre-operative evidence, presenting all pre-operative imaging and laboratory evidence verbatim. The second was the conclusion-masked version, which cleared the conclusion field across all 873 imaging reports and replaced 124 direct diagnostic mentions across 118 reports in 113 cases with "liver malignancy" while retaining objective imaging findings; the eight replaced diagnostic terms commonly found in the original Chinese medical records corresponded to "hepatocellular carcinoma", "intrahepatic cholangiocarcinoma", "cholangiocarcinoma", "primary liver cancer", "liver cancer", "HCC", "ICC", and "hepatic Ca", a local clinical shorthand for liver carcinoma. The third was the imaging-free version, which removed all 873 imaging reports and retained only age, sex, and the 28,304 laboratory test records.

Within the 181-case in-hospital liver primary cohort, 5 cases had no structured imaging reports and 1 case had no laboratory records retrieved within the 90-day window; within the 514-case combined enriched cohort, 11 cases lacked imaging reports in the window. All cases with partial missingness were retained in the analysis sets, and none entered the top-50% accepted group. The case-level input inventory indexed by de-identified study ID for the 181-case in-hospital liver primary cohort is provided as Supplementary Data 1, completeness checks for the primary cohort are detailed in Supplementary Note S2, and verification results for the 514-case combined enriched cohort are provided in Supplementary Note S11.

### Local clinical AI agent architecture and run configuration

All primary local experiments were deployed on a single-GPU NVIDIA GB10 DGX Spark inference platform running the open-source Alibaba Cloud Qwen3.5-4B instruction-tuned model without task-specific fine-tuning via vLLM version 0.31.0 at 16-way concurrency (Fig. 1a). Across all 4B primary experiments, the same local model served sequentially in four roles—patient simulator, consulting doctor model, reader model, and local 4B judge—without weight fine-tuning or training an auxiliary reliability estimator. Default sampling temperatures were 0.6 for the consulting doctor model, 0.01 for the patient simulator, and 0 for the reader model. Each cohort was evaluated under the single-run tier followed by five independent repeated runs to construct five-run consistency and the five-run consistency tier; complete prompt templates and environment/hardware specifications are provided in the supplementary material (Supplementary Notes S5 and S6).

Across the three multi-turn interactive benchmarks—MIRA-v2, CDM, and VivaBench—the patient simulator responded solely from source record background, including chief complaint, history of present illness, and pre-admission medications, without invoking external tools, whereas the doctor model acquired physical examination findings, laboratory tests, and imaging studies through multi-turn natural-language history taking and structured tool calls. Each consultation was capped at 10 interaction rounds, and system code enforced at least 3 dialogue turns and at least 2 non-submission laboratory or imaging tool calls before permitting final diagnosis submission. In MIRA-v2 and CDM, owing to the absence of local retrieval indices, surgical procedure tools were removed while medication-prescription tools were retained; the 4B doctor model invoked the prescription tool in 92.74% and 94.66% of cases during the first independent run, respectively, whereas the original VivaBench toolset contained no intervention tools. Accuracy judging across all experiments evaluated only the final diagnosis submitted by the doctor model.

Within the 10-round cap of the first run, case termination fell into three categories: normal tool-call submission, system-forced termination at the 10-round limit, and incomplete runs lacking an end marker. Across the 551 MIRA-v2 cases, the 4B doctor model achieved 86.03% normal submissions, 13.97% forced terminations, and 0% incomplete runs; across the 599 CDM cases, the proportions were 93.49%, 6.51%, and 0%. Across the 274 VivaBench cases in the three-doctor shared set, the proportions of normal submission, forced termination, and abnormal termination were 93.43%, 5.84%, and 0.73% for the 4B doctor model; 97.81%, 0%, and 2.19% for the 27B doctor model; and 96.35%, 3.65%, and 0% for cloud DeepSeek V4.1 Flash.

For single-step static record reasoning in the in-hospital liver cohort, the doctor model received the complete pre-operative evidence text from one of the three input versions; we removed the abstention instruction allowing the model to state insufficient information or uncertainty, provided no candidate diagnostic options, and required a two-line free-text output giving the single most likely diagnosis followed by a one-sentence clinical rationale, ensuring every case yielded a definite diagnosis without uncertain or invalid categories. Simultaneously, the identical objective clinical evidence was passed independently to the reader model configured with the evidence sufficiency assessment prompt, without passing the doctor model's diagnosis or rationale (Supplementary Note S5).

Cross-model substitution experiments kept system prompts, submission constraints, and judging criteria identical and comprised two orthogonal control arms. In the consulting-doctor substitution arm with the local Qwen3.5-4B reader model held fixed, we replaced both the consulting doctor model and its paired patient simulator in multi-turn simulations on the 274-case VivaBench three-doctor shared set with the larger 27B doctor model or cloud DeepSeek V4.1 Flash, evaluating the objective clinical evidence actually collected by each doctor model (Supplementary Note S8.2); on the complete version of usable pre-operative evidence in the 514-case combined enriched cohort, we held the pre-operative objective evidence inputs identical and replaced only the single-step consulting doctor model with the 27B doctor model (Supplementary Note S11.2).

In the reader-model substitution arm, existing consultation trajectories were held fixed and only the reader model was replaced with Qwen3.5-27B or Google DeepMind Gemma 4 E4B. We first compared ranking discrimination across reader models on the 524-case early paired development subset of MIRA-v2 with fixed 4B doctor model trajectories (Supplementary Fig. S1), and subsequently evaluated evidence sufficiency score distributions on 381 MIRA-v2 cases with fixed 4B doctor model trajectories and the 274-case VivaBench three-doctor shared set with fixed 27B doctor model trajectories to examine whether larger or cross-family reader models exhibited score compression or loss of discrimination (Supplementary Note S8.1).

### Evidence sufficiency assessment and comparator reliability signals

When the consulting doctor model is ready to conclude a multi-turn consultation or complete single-step record analysis and submit a diagnosis, the system performs a decision-point evidence sufficiency assessment (Fig. 1b). This assessment is built on AnyJev[17] version 0.3.0, a third-party open-source implementation of the structured decision readout interface JEV (https://typesafe.ai); all reliability signals and their parameter definitions are summarized in Supplementary Table S2.

Submission-point forward readout followed three strict design constraints. The first was objective-evidence isolation: the assessment input retained only collected objective clinical evidence—chief complaint, history-taking dialogue, physical examination, laboratory tests, and imaging studies—while completely removing the consulting doctor model's intermediate reasoning process and proposed diagnosis, thereby preventing plausible-sounding reasoning narratives from distorting evidence sufficiency assessment (Supplementary Fig. S9c). The second was fixed-order forward readout at temperature 0: with temperature set to 0 and no free-text generation, the model read the output logits of option-index tokens in fixed monotonically increasing order from "Not sufficient at all" to "Fully sufficient" without option-position permutation so as to preserve the ordinal scale structure (Supplementary Note S5). The third was prior debiasing and expected scoring: using three content-free neutral placeholder texts—empty text, "N/A", and "[MASK]"—we measured and cached the mean baseline logits $\bar{\ell}_i^{(0)}$ across the five levels for batch reuse, subtracted this prior baseline in logit space before normalizing, and computed the weighted expectation using equidistant weights $w_i \in \{0, 0.25, 0.50, 0.75, 1.00\}$:

$$S_{\mathrm{ES}} = \sum_{i=1}^{5} w_i\, \tilde{p}_i, \qquad \tilde{p}_i = \frac{\exp\big(\ell_i - \bar{\ell}_i^{(0)}\big)}{\sum_{j=1}^{5} \exp\big(\ell_j - \bar{\ell}_j^{(0)}\big)}$$

where $\ell_i$ denotes the raw output logits of the $i$-th option-index token. During the research phase, we also recorded an evidence sufficiency assessment after every tool call yielding new evidence—termed per-evidence readout—executing 8,033 readouts across all 551 MIRA-v2 cases, with a mean of 14.58 per case, to track dynamic evolution throughout consultation; in primary analyses and actual clinical deployment, only 1 forward readout was executed per case at the diagnostic submission point (Supplementary Fig. S8). Notably, given identical fixed objective clinical evidence, fixed-order forward readout at temperature 0 is completely deterministic with zero sampling variance; across five independent runs of a case in multi-turn benchmarks, differences in evidence sufficiency scores arose solely from dialogue trajectories and evidence collected under the consulting doctor model's sampling temperature of 0.6, whereas in the static single-step in-hospital liver cohort, because the three input texts remained fixed, five-run repetition resampled only the doctor model's diagnostic output and required only 1 forward readout per fixed input text.

Comparator reliability signals were grouped into passive reliability signals extracted post-consultation and controlled comparators read at the diagnostic submission point (Supplementary Fig. S9). Passive reliability signals comprised token generation probability and five-run consistency. Token generation probability took the geometric mean of predicted token probabilities across the final diagnosis sequence without extra model calls; because 1 MIRA-v2 trajectory and 6 CDM trajectories lacked saved token probabilities, valid analysis sets for this signal were 550 and 593 cases, respectively. Five-run consistency was computed according to task format: across the three multi-turn benchmarks, the five final diagnostic texts for each case were encoded into sentence embeddings via `all-MiniLM-L6-v2` and averaged across pairwise cosine similarities clamped to 0–1; in the in-hospital liver cohort, first-line diagnoses from the five independent runs were mapped to predefined categories via a dictionary, taking the proportion of the most frequent category.

Controlled comparators at the diagnostic submission point comprised four single-run methods designed to disentangle scale granularity from objective-evidence isolation. Direct 0–1 probability reporting used the identical objective-evidence-only prompt as evidence sufficiency assessment but prompted the model to generate a numerical probability between 0 and 1 as free text. The remaining three comparators supplied the doctor model's submitted diagnosis alongside objective clinical evidence: self-reported diagnostic confidence prompted the model to output a 0–1 confidence number as free text, whereas binary sufficiency assessment over "Yes" versus "No" options and ten-level sufficiency assessment over a 0–10 scale applied the same temperature-0 output-logit readout and content-free prior debiasing as evidence sufficiency assessment.

## Diagnostic accuracy judging, single final diagnosis merging, and error attribution

Diagnostic correctness across all four clinical cohorts was determined by the local Qwen3.5-4B judge following standardized rules, with complete judging prompts provided in the supplementary material (Table 1 and Supplementary Note S5). Across all 551 MIRA-v2 cases, the local 4B judge evaluated each submitted diagnosis against the reference diagnosis, and cloud DeepSeek V4.1 Flash performed full-set independent re-judging to verify judging agreement; local judging was subsequently adopted uniformly in primary analyses (Table 1 and Supplementary Fig. S11). Across all 599 CDM cases, the local 4B judge evaluated submissions against full International Classification of Diseases diagnoses, with a stratified sample of 25 cases per condition, totaling 100 cases, re-judged by the cloud model; cross-model re-judging agreement across all public benchmarks is summarized in Table 1 and Supplementary Note S9.

On the 274-case VivaBench three-doctor shared set, the local 4B judge applied the judging prompt from the upstream evaluation framework[11] to implement the top-1 approximate criterion[38], which credits hitting either the primary diagnosis or an acceptable differential diagnosis, with independent cloud re-judging stratified across the three consulting doctor models (Table 1 and Supplementary Note S9). In the 181-case in-hospital liver primary cohort and 514-case combined enriched cohort, reference labels were derived from pathology-confirmed diagnoses and verified through machine-assisted rule checks without case-by-case independent blinded clinician re-review. The local 4B judge determined whether the first-line free-text diagnosis matched the pathology category of hepatocellular carcinoma or intrahepatic cholangiocarcinoma, alongside checks of exact first-line literal matching and agreement with predefined keyword rules (Table 1 and Supplementary Table S3).

The single-run tier was scored on the final diagnosis submitted in the first independent run on public multi-turn benchmarks and on the single-run first-line free-text diagnosis against pathology reference labels in the in-hospital liver cohort. The five-run consistency tier aggregated five independent outputs using a single final diagnosis by consensus rule without accessing reference labels. Specifically, on MIRA-v2, CDM, and VivaBench, the five final diagnostic texts were encoded into sentence embeddings via `all-MiniLM-L6-v2` and clustered via single-linkage connected components at a pairwise cosine similarity threshold of at least 0.80; when the largest cluster contained at least 2 members, the diagnosis within that cluster having the highest sum of pairwise cosine similarities to all other cluster members was selected as the single final diagnosis. Ties between largest clusters were resolved deterministically by higher sum of within-cluster cosine similarities followed by earlier run index across runs 1–5; when the largest cluster contained fewer than 2 members, indicating no consensus, no automated diagnosis was output and the case was deferred directly to clinician review.

In the in-hospital liver cohort, first-line diagnoses from the five free-text outputs were mapped via a predefined keyword dictionary to four categories—"hepatocellular carcinoma", "intrahepatic cholangiocarcinoma", "other", and "indeterminate"—selecting the most frequent category as the single final diagnosis, with ties broken deterministically by predefined category order. All merged diagnoses entered the accepted-group evaluation, with outputs mapped to "other", "indeterminate", or mismatching the pathology reference label counted as errors; across all three input versions of the 181-case in-hospital liver primary cohort, the modal category appeared at least 3 times in every case. "Correct in at least three of five runs" was reported solely as an auxiliary criterion in overlap sensitivity analyses (Supplementary Note S10), and robustness checks across similarity thresholds of 0.70–0.90 and complete-linkage clustering are detailed in Supplementary Note S1.4.

Clinical mechanism attribution for failure cases and typical-case verification were performed case by case by a single investigator using prespecified coding rules. In CDM, all 111 single-run misdiagnoses were categorized by the clinical relationship between the first-run diagnosis and the reference diagnosis into three classes: broader acute abdominal-pain spectrum outside the four target diseases, misclassification into another target acute abdominal disease, and wording or granularity differences within the same disease (Fig. 3b), alongside disease-stratified accuracy reporting (Supplementary Table S1). Additionally, for representative CDM cases with high evidence sufficiency scores that were misdiagnosed in the first run but correct in most remaining runs, the investigator compared laboratory and imaging evidence collected and diagnostic rationales across all five independent runs.

In the 225-case five-run complete subset of VivaBench, the investigator reviewed first-run diagnostic rationales for all 31 cases misdiagnosed in all five runs and coded three indicators: whether the reference disease was mentioned, whether differential diagnoses were listed, and whether a definitive diagnosis was made despite missing key confirmatory tests. Diagnostic deviations were then classified into three types: treating clinical syndromes or symptoms as final diagnoses, selecting the wrong specific entity within the same disease family, and over-relying on unrelated nonspecific signs, while counting exact literal distinct diagnostic answers per case across the five independent runs.

### Triage evaluation metrics and compute cost accounting

We defined three case-level triage evaluation metrics. Full-cohort single-run accuracy (also termed single-run consultation accuracy on multi-turn benchmarks) denotes the proportion of correctly diagnosed cases across all evaluated cases in a single run without triage; auto-accept fraction, also termed coverage, denotes the proportion of all cohort cases routed to the auto-accept branch by descending reliability score, excluding non-consensus cases deferred to clinician review in the five-run consistency tier so as to reflect cases genuinely spared clinician review; and pass rate among accepted cases denotes the proportion of auto-accepted cases with a correct final diagnosis. For a given high-safety target, maximum auto-accept fraction was defined as the largest proportion of cases that could be auto-accepted while maintaining a pass rate among accepted cases at or above the target line. When plotting descriptive triage curves within a single cohort, cases were ranked by descending reliability score and cut off at the rounded $qN$-th case; when a threshold fell inside a tied group, tie order was randomly permuted 2,000 times to report the mean and 2.5%–97.5% quantile interval, with exact auto-accept fractions detailed in Supplementary Table S6 for MIRA-v2 and Supplementary Note S8.3 for CDM. Unlike these single-cohort descriptive analyses, deployment simulations did not rely on batch-level ranks and instead classified each incoming case independently against the frozen per-case threshold locked on the calibration set.

Compute cost efficiency was measured by complete consultation runs per correctly accepted case:

$$\text{Runs per correctly accepted case} = \frac{k}{\text{Auto-accept fraction} \times \text{Pass rate among accepted cases}} = \frac{kN}{\text{Correctly accepted cases}}$$

where $k$ is the number of complete consultation runs per case, with $k$ = 1 for the single-run tier and $k$ = 5 for the five-run consistency tier, and $N$ is total cohort size (Fig. 2c). Evidence sufficiency assessment required a forward pass over approximately 10,000 context tokens, whose additional compute overhead was converted into full consultation-equivalents using measured wall-clock runtime. Across all 551 MIRA-v2 cases, serial completion of one full consultation required 36.31 h, or ~237 s per case, whereas executing 8,033 per-evidence readouts during the research phase required 9.68 h, giving a ratio of 0.27. For conservative cost accounting, main results tables adopted an upper bound incorporating research-phase exploratory overhead, counting 1.27 consultation-equivalents for the single-run tier, 5.27 for the five-run consistency tier, and 0 for the full-deferral tier (Table 2).

In real-world clinical deployment, only one fixed-order forward probability readout is executed per case at the final diagnostic submission point. Under 16-way concurrency across all 551 MIRA-v2 cases, total submission-point forward readout runtime was 6.1 min, corresponding to an amortized server compute time of ~0.7 s per case, with a mean client-side request wait time including queuing of 10.4 s and a range of 0.5–25.9 s, accounted as 0.02 full consultation-equivalents—corresponding to 1.02 and 5.02 consultation-equivalents for the single-run tier and five-run consistency tier, respectively, under actual deployment accounting. In the static single-step in-hospital liver cohort, mean runtime per case was 1.54 s for a single submission-point forward readout and 1.61 s for a single diagnostic inference. All compute accounting captured only on-premise local model inference overhead and excluded clinician review time.

### Three-tier sequential triage protocol and local statistical certification

Drawing on selective classification and distribution-free risk-controlling prediction sets[44,45], the protocol locks three governance parameters prior to deployment according to the hospital's risk tolerance, avoiding a backward search on the calibration set for maximum coverage at the target line. First, the nominal auto-accept fraction $q$ was fixed at 50% for the primary evaluation, with 20% and 80% examined in sensitivity analyses. Second, the clinical target line $T$—defined as the minimum pass rate among accepted cases that auto-accepted cases must achieve—was uniformly prespecified at 95% across all cohorts without dataset-specific adjustment; because no configuration on VivaBench met this standard, 90% was evaluated as a relaxed admission floor in sensitivity checks, whereas raising $T$ to 98% and 98.9% was used solely for high-safety-target sensitivity analyses. Third, the one-sided confidence parameter *z* was set to *z* = 1.6449, corresponding to 95% one-sided confidence, for the primary criterion, alongside *z* = 1.0, corresponding to approximately 84% one-sided confidence, as a lenient methodological boundary comparator (Fig. 4a).

The protocol evaluates three tiers in ascending order of computational cost. The first is the single-run tier, which executes one independent consultation or single-step record inference per patient followed by evidence sufficiency assessment. The second is the five-run consistency tier, intended for workflows in which patient history and investigation results have already been compiled and multiple back-end model inference calls are permitted: the same consulting doctor model independently runs five times, cases are ranked by five-run consistency, and submissions are scored using a single final diagnosis by consensus; by contrast, in real-time outpatient consultation settings with strict constraints on response latency and local concurrent compute, the system should not incur fivefold inference cost and instead defers directly to clinician review whenever the single-run tier fails certification. If neither the first nor second tier passes certification, the protocol enters the full-deferral tier—halting auto-acceptance, deferring all cases to clinician review, and initiating a re-calibration and re-certification workflow through consulting doctor model upgrading or lowering the nominal auto-accept fraction *q*. In addition, a within-model two-stage cascade using the same 4B model failed to reach the prespecified admission standard relative to the five-run consistency tier on the multi-turn CDM cohort and yielded a pass-rate gain of at most 0.14 percentage points on the static in-hospital liver cohort; it was therefore eliminated during pre-evaluation (Supplementary Note S8.3).

During live triage, decisions do not depend on the relative ranks of a batch of patients and are instead adjudicated case by case against a frozen per-case threshold locked on the local calibration set. Specifically, on a calibration set of size $N_{\mathrm{cal}}$, all cases are sorted in descending order by their primary reliability score and broken by a prespecified secondary sort key in descending order, taking the primary-score and secondary-key pair of the $\lceil qN_{\mathrm{cal}} \rceil$-th ranked case as the frozen threshold. For the single-run tier, the evidence sufficiency score itself serves as the primary reliability score, and the secondary sort key is identical to it; for the five-run consistency tier on multi-turn benchmarks, five-run consistency serves as the primary reliability score, and the mean evidence sufficiency score across the five independent runs—which is invariant to run order—serves as the secondary sort key (Supplementary Note S1.3). After deployment, each incoming patient is evaluated independently at the diagnostic submission point: cases whose primary reliability score exceeds the frozen primary score, or equals the frozen primary score with a secondary sort key at or above the frozen secondary key, are auto-accepted; all remaining cases, as well as cases lacking consensus in the five-run consistency tier, are deferred to clinician review. Because this rule triggers independently on a per-case basis without relying on batch size, waiting queues, or case IDs, the realized auto-accept fraction on future test sets fluctuates naturally around the nominal fraction *q* (Supplementary Note S1.3).

Admission certification is executed within the calibration accepted group. For the five-run consistency tier on public multi-turn benchmarks, cases deferred to clinician review owing to lack of consensus are excluded first; letting *m* denote the number of accepted cases and *c* the number of correctly diagnosed cases in the calibration accepted group, the observed pass rate among accepted cases is $\hat{p} = c/m$, from which the one-sided Wilson lower confidence bound on the true pass rate is computed as:

$$\mathrm{LCL}_z = \frac{\hat{p} + \frac{z^2}{2m} - z\sqrt{\frac{\hat{p}(1-\hat{p})}{m} + \frac{z^2}{4m^2}}}{1 + \frac{z^2}{m}}$$

A tier passes calibration certification if and only if *m* > 0 and $\mathrm{LCL}_z \geq T$. Under zero misdiagnoses in the accepted group, the lower bound simplifies to $m/(m+z^2)$: satisfying *T* = 95% requires at least 52 accepted cases with zero errors under the primary 95% one-sided criterion (*z* = 1.6449) and at least 19 error-free cases under the lenient comparator (*z* = 1.0); in a 60-case calibration accepted group, a single error lowers the primary lower bound to 92.9% and triggers rejection. Furthermore, sequentially testing the single-run tier and, upon failure, the five-run consistency tier on the same calibration set yields a theoretical upper bound on the overall type I error rate of 2*α*; under a conservative Bonferroni-corrected comparator tightening the per-step significance level to *α*/2 = 0.025 (*z* = 1.9600), at least 73 error-free accepted cases are required for certification, with sensitivity analyses detailed in the supplementary material (Supplementary Note S1.2).

To evaluate sampling stability, we performed 200 complementary random splits without replacement into calibration and test sets, stratified by clinical condition, specialty group, or pathology category. Calibration set sizes $N_{\mathrm{cal}}$ were set to 300, 300, 150, and 120 cases for MIRA-v2, CDM, the 274-case VivaBench three-doctor shared set, and the in-hospital liver primary cohort, respectively, with the remaining cases forming the corresponding independent test sets; for VivaBench, the five-run consistency tier was re-split into calibration and test sets on the 225-case five-run complete subset. All 200 random splits, as well as all bootstrap resamplings and permutation tests in this study, were computed offline directly from frozen saved consultation trajectories, diagnostic outputs, and reliability score tables without re-invoking large language model inference.

To strictly separate tier selection from independent held-out evaluation, protocol performance was partitioned into four evaluation objects. The first and second objects—the candidate single-run tier and the candidate five-run consistency tier—were each evaluated independently across the 200 random splits using three metrics: calibration certification pass rate, test-set attainment rate among certified splits, and all-split test attainment rate, treating uncertified splits as non-attaining. The third, primary evaluation object was the actual sequential policy: on each split, the system decided solely from the calibration set—deploying the single-run tier if certified, testing the five-run consistency tier if not, and entering the full-deferral tier if neither passed—and reported on the independent test set the all-split test attainment rate, calibration false-certification rate, full-deferral rate, and mean consultation-equivalents per case; here, calibration false-certification rate denotes the proportion of splits certified on the calibration set whose test-set pass rate among accepted cases fell below *T*, and the first three proportions sum strictly to 100% (Table 2). The fourth object was an exploratory post hoc cohort recommendation aggregated across all 200 simulated splits for retrospective reference only and not used for pre-deployment selection; simulation outcomes under the lenient comparator and mutually exclusive split breakdowns are provided in Supplementary Note S1.3. Additional sensitivity analyses examining backward search for maximum coverage or score thresholds on the calibration set are given in Supplementary Note S1.1, and analyses across calibration sample sizes from 50 to 400 cases are provided in Supplementary Note S1.2.

### Statistics and reproducibility

Because reliability scores were markedly skewed and several calibration accepted groups and clinical subgroups had small sample sizes—such as $m \leq 60$ in the accepted group of the in-hospital liver primary cohort, $n = 64$ in the CDM colonic diverticulitis subgroup, and $n = 25$ in the in-hospital liver primary cohort intrahepatic cholangiocarcinoma subgroup—and because observed pass rates among accepted cases approached 100% or reached zero errors at several auto-accept fractions, conventional Wald normal-approximation intervals degenerate to zero width when $\hat{p} = 1$ and risk under-coverage or boundary overshoot outside 0–1. We therefore used the Wilson score interval for uncertainty estimation and admission certification of all binomial proportions, combined with prespecified non-parametric resampling methods for statistical inference.

Across all clinical cohorts, valid subsets, and the combined enriched cohort, two-sided 95% Wilson confidence intervals were reported for both pass rates among accepted cases at fixed auto-accept fractions and full-cohort single-run accuracies, with item-by-item intervals listed in Supplementary Note S7.1. Condition-specific subgroup confidence intervals across the seven acute conditions in MIRA-v2 are presented in Supplementary Fig. S7.

Pre-deployment admission certification applied one-sided Wilson lower confidence bound tests within each calibration accepted group: at the nominal 50% auto-accept fraction, calibration accepted-group sample sizes were $m = 150$ for MIRA-v2 and CDM, $m = 75$ for VivaBench, and $m = 60$ for the in-hospital liver primary cohort, each tested against the prespecified clinical target line *T* under both the primary and lenient confidence criteria (Table 2). In full-range ranking ablation comparisons reported in the supplementary material, discrimination was quantified by the area under the receiver operating characteristic curve (AUC), defined as the probability that a randomly drawn correctly judged case receives a higher reliability score than a randomly drawn incorrectly judged case, with ties counted as 0.5.

Within-cohort paired comparisons used 2,000 case-level paired bootstrap resamples to compute two-sided 95% percentile confidence intervals for differences and two-sided exact *P* values. On the 524-case development subset of MIRA-v2, paired bootstrapping tested AUC differences across reader models, yielding $P = 0.857$ for Qwen3.5-27B versus Qwen3.5-4B and $P = 0.302$ for Gemma 4 E4B versus Qwen3.5-4B (Supplementary Fig. S1b). Across all 551 MIRA-v2 cases, paired bootstrapping compared the AUC of the research-phase per-evidence readout endpoint against the single deployment forward readout (Supplementary Fig. S8b), and the same method evaluated 30 paired signal comparisons (Supplementary Note S7.2). On the 274-case VivaBench three-doctor shared set, paired bootstrapping compared top-50% pass rates among accepted cases across reader models (Supplementary Note S8.1); on the 181-case in-hospital liver primary cohort, 333-case newly added enrichment set, and 514-case combined enriched cohort, it tested full-cohort single-run accuracy differences between doctor models and the simple rule-based baseline searching only for cancer keywords in imaging reports (Supplementary Table S3).

To quantify coverage fluctuations arising from high-score ties in direct 0–1 probability reporting, we used 2,000 random permutations within tied groups to estimate means and 2.5%–97.5% quantile intervals on the 551 MIRA-v2 cases and 599 CDM cases (Supplementary Table S6 and Supplementary Note S8.3). On the 181-case in-hospital liver primary cohort, 333-case newly added enrichment set, and 514-case combined enriched cohort, superiority of the top-50% pass rate among accepted cases over random triage for each input version was assessed via 2,000 one-sided permutation tests, yielding exact one-sided *P* values on the 181-case primary cohort of $P = 0.0005$ for the complete version of usable pre-operative evidence, $P = 0.0030$ for the conclusion-masked version, and $P = 0.0480$ for the imaging-free version, with exact *P* values for all cohorts and input versions detailed in Supplementary Table S3. Uncertainty intervals for full-range single-signal ranking metrics and sensitivity analyses across 50–400 calibration sample sizes and backward threshold search were estimated using 1,000 bootstrap or stratified resamples (Supplementary Figs. S1a, S4b, and S9a; Supplementary Notes S1.1 and S1.2). The intrahepatic cholangiocarcinoma prevalence-gradient curves at fixed $N = 333$ spanned 12 target gradients from 5.0% to 49.0%, constructing 10 stratified simulated cohorts without replacement per gradient from the 514-case complete version to evaluate top-50% single-run tier pass rates among accepted cases, followed by 200 stratified random calibration–test splits with $N_{\mathrm{cal}} = 120$ per simulated cohort to evaluate calibration certification pass rates (Fig. 5b, c and Supplementary Note S11.1).

Except for the one-sided Wilson lower confidence bound tests for admission certification and the one-sided permutation tests in the in-hospital liver cohort, all hypothesis tests were two-sided with a significance level of $\alpha = 0.05$, and all exact *P* values are reported individually in the corresponding main figure legends, supplementary figure legends, and supplementary tables. For the 30 paired comparisons in the supplementary material comprising 5 signal contrasts across 3 evaluation metrics and 2 judging criteria, the Holm step-down procedure controlled the family-wise type I error rate under three family definitions: 10-test families stratified by evaluation metric, 5-test families stratified by metric and judging criterion, and a single conservative 30-test family; all adjusted exact $P$ values across all three family definitions satisfied $P \geq 0.110$ (Supplementary Note S7.2). Multiplicity sensitivity analysis for the two-step sequential admission certification used the Bonferroni-corrected comparator described above (Supplementary Note S1.2).

No formal a priori power calculation was performed; cohort sample sizes were determined jointly by available public benchmark sizes, prespecified stratified sampling counts, and eligible pathology-confirmed cases within the hospital ethics approval window. Aside from cases lacking five complete runs or used for preliminary system debugging as specified in Study design and clinical cohorts, no additional data exclusions were applied, and all simulations and resamplings used fixed random seeds for full reproducibility. Server-side primary analyses ran on Python 3.12.3, PyTorch 2.14.1, and transformers 5.19.0; local deployment certification simulations and plotting ran on Python 3.11.5 with NumPy 1.24.3, pandas 2.0.3, SciPy 1.11.1, scikit-learn 1.7.0, and Matplotlib 3.7.2 (Supplementary Note S6).

### Ethics approval and data governance

This study adhered to the Declaration of Helsinki; all analyzed data comprised de-identified or pseudonymized retrospective clinical records, and no participants were prospectively recruited. MIRA-v2 and CDM were reconstructed under the PhysioNet credentialed access agreement from the de-identified retrospective databases MIMIC-IV v2.2 and MIMIC-IV-Note v2.2, with MIRA-v2 additionally incorporating MIMIC-IV-ED v2.2[32–35]. Collection and release of MIMIC-IV resources were approved by the institutional review board of Beth Israel Deaconess Medical Center with a waiver of informed consent; all data analysts completed the Collaborative Institutional Training Initiative (CITI) "Data or Specimens Only Research" training and complied strictly with the PhysioNet Credentialed Health Data Use Agreement 1.5.0.

VivaBench is a publicly released de-identified clinical case benchmark distributed under the Creative Commons Attribution 4.0 International (CC BY 4.0) license[38]. The in-hospital liver malignancy primary cohort and combined enriched cohort constituted a retrospective analysis of de-identified medical records approved by the Biomedical Research Ethics Committee of West China Hospital, Sichuan University, approval number 2024(189), with a waiver of informed consent; the methods and analysis status for the benchmarks and 181-case primary cohort were retrospectively registered on the Open Science Framework on 12 November 2026 under DOI 10.17605/OSF.IO/S5DMH, with the additional 333 cases accrued afterward under identical criteria for subtype-prevalence sensitivity analyses.

Before exporting in-hospital liver cohort inputs for model evaluation, the research team removed all direct personal identifiers including patient names, hospital admission numbers, and exact calendar dates; model inputs contained only the de-identified clinical fields specified in Study design and clinical cohorts, without raw pathology reports or reference labels. Source medical records containing patient identifiers remained strictly within the hospital's controlled physical environment and were never transmitted off-site. All consulting doctor model inference, evidence sufficiency assessment, automated judging, and deployment certification simulations for the private in-hospital liver cohort were executed offline on local hospital servers using local models, without transmitting protected records over external networks or retaining logs or retraining on any public artificial intelligence service.

### Code availability

The custom code used for multi-turn interactive consultation, evidence sufficiency assessment, three-tier sequential triage protocol simulations, and statistical analyses—together with complete prompt templates and 200-split simulation details releasable under respective data use agreements—will be made publicly available upon publication via GitHub (https://github.com/BreezeHavana/reliable-jev-paper); access during peer review is available upon request to the editors and reviewers. Complete verbatim prompt templates are provided in Supplementary Note S5. In addition, all prompt templates and deterministic text-rendering code were frozen on 12 November 2026 prior to formal evaluation; six SHA-256 checksums covering the three information-ablation input files and their deterministically rendered texts for the in-hospital liver malignancy primary cohort are archived with run logs and listed in full in Supplementary Note S2 for reproducibility verification.

### Use of generative AI

During manuscript drafting and language revision, the authors used the generative artificial intelligence tools Gemini and DeepSeek V4.1 Flash to improve readability, refine paragraph organization, and polish phrasing. These tools were used exclusively for language and editorial refinement and were not used to formulate hypotheses, generate experimental data, perform statistical analyses, or compile references; no protected patient source records or unpublished sensitive data were uploaded to any public or insecure artificial intelligence system. All authors manually verified and edited all scientific arguments, statistical values, figures, tables, and references and take full responsibility for the published content; the experimental use of DeepSeek V4.1 Flash as a comparator consulting doctor model and independent re-judge is detailed in earlier sections. Additional details on study design, data inclusion, and statistical verification are provided in the accompanying Nature Portfolio Reporting Summary.

## Data availability

Public and restricted datasets supporting the findings of this study are accessible as follows. The public datasets MIMIC-IV v2.2 (https://doi.org/10.13026/6mm1-ek67), MIMIC-IV-ED v2.2 (https://doi.org/10.13026/5ntk-km72), and MIMIC-IV-Note v2.2 (https://doi.org/10.13026/1n74-ne17) are available to credentialed researchers who complete CITI training and sign the PhysioNet Credentialed Health Data Use Agreement 1.5.0. MIRA-v2 can be reconstructed from MIMIC-IV v2.2 using the developers' public repository (https://github.com/Dyke-F/MIRA); CDM can be reconstructed from the developers' public repository (https://github.com/paulhager/MIMIC-Clinical-Decision-Making-Dataset) or obtained directly from PhysioNet (https://physionet.org/content/mimic-iv-ext-cdm/1.0/). Owing to PhysioNet data use restrictions, derived cohorts and case-level records containing participant data are not publicly released. VivaBench is publicly available under the Creative Commons Attribution 4.0 International license (https://huggingface.co/datasets/chychiu/VivaBench). The in-hospital liver malignancy primary cohort and combined enriched cohort contain individual-level clinical information and are not publicly available under West China Hospital data governance policies and ethics approval; de-identified research data may be requested from the corresponding author upon reasonable request, subject to approval by the Biomedical Research Ethics Committee of West China Hospital, Sichuan University and execution of a data use agreement. The case-level input inventory indexed by study ID for the 181-case in-hospital liver primary cohort is provided with this manuscript as Supplementary Data 1, detailing study ID, pathology reference category, evidence item counts across the three input versions, missing-value counts, visible evidence day ranges relative to report dates, and curation citation flags, without direct personal identifiers or raw pathology text; the crosswalk linking study IDs to hospital source record numbers is withheld under ethics approval and hospital data governance rules.

## Acknowledgements

This study was funded by the National Natural Science Foundation of China, grant 82400708, and the Science and Technology Department of Sichuan Province, grant 2025ZNSFSC1683. The funders played no role in study design, data collection, analysis and interpretation of data, or the writing of this manuscript.

## Author contributions

Y.Q., C.S., S.S. and W.W. conceived and designed the study. Y.Q. and C.S. contributed equally to this work. Y.Q. constructed the in-hospital clinical cohort and performed retrospective information masking, de-identification and clinical review. C.S. conducted the local clinical AI agent experiments, statistical analyses, three-tier sequential triage protocol deployment simulations and figure generation. S.S. performed clinical mechanism attribution of failure cases and pathology diagnosis verification. W.W. guided and supervised study implementation and clinical interpretation. Y.Q. and C.S. co-wrote the initial draft of the manuscript, and S.S. and W.W. contributed to revising key scientific content. All authors read, verified and approved the final manuscript.

## Competing interests

Che Sun is employed by Shenzhen Mindray Bio-Medical Electronics Co., Ltd., which did not fund this study and had no role in study design, data analysis and interpretation, or the decision to publish. Apart from this employment, all authors declare no financial or non-financial competing interests.

---